\documentclass[final,3p,times]{elsarticle}

\usepackage{graphicx}%
\usepackage{multirow}%
\usepackage{amsmath,amssymb,amsfonts}%
\usepackage{amsthm}%
\usepackage{mathrsfs}%
\usepackage{booktabs}%
\usepackage{subfigure}
\usepackage{xcolor}
\usepackage{xurl}
\usepackage{footnote}

\journal{Engineering Applications of Artificial Intelligence}

\begin{document}

\begin{frontmatter}



\title{GAMMA-Net: Adaptive Long-Horizon Traffic Spatio-Temporal Forecasting Model based on Interleaved Graph Attention and Multi-Axis Mamba}

\author{Dongyi He}
\author{Yuanquan Gao}
\author{He Yan}
\author{Bin Jiang\corref{cor1}} 
\cortext[cor1]{Corresponding author. Email: jb20200132@cqut.edu.cn}

\affiliation{organization={School of Artificial Intelligence, Chongqing University of Technology},
            city={Chongqing},
            postcode={400054}, 
            country={China}}

\begin{abstract}
  Accurate traffic forecasting is crucial for intelligent transportation systems, supporting effective traffic management, congestion reduction, and informed urban planning. However, traditional models often fail to adequately capture the intricate spatio-temporal dependencies present in traffic data. To overcome these limitations, we introduce GAMMA-Net, a novel approach that integrates Graph Attention Networks (GAT) with multi-axis Selective State Space Models (Mamba). The GAT component uses a self-attention mechanism to dynamically adjust the influence of nodes within the traffic network, enabling adaptive spatial dependency modeling based on real-time conditions. Simultaneously, the Mamba module efficiently models long-term temporal and spatial dynamics without the heavy computational cost of conventional recurrent architectures. Extensive experiments on several benchmark traffic datasets, including METR-LA, PEMS-BAY, PEMS03, PEMS04, PEMS07, and PEMS08, show that GAMMA-Net consistently outperforms existing state-of-the-art models across different prediction horizons, achieving up to a 16.25\% reduction in Mean Absolute Error (MAE) compared to baseline models. Ablation studies highlight the critical contributions of both the spatial and temporal components, emphasizing their complementary role in improving prediction accuracy. In conclusion, the GAMMA-Net model sets a new standard in traffic forecasting, offering a powerful tool for next-generation traffic management and urban planning. The code for this study is available at \url{ https://github.com/hdy6438/GAMMA-Net}
\end{abstract}




\end{frontmatter}

\section{Introduction}
\label{sec:intro}
Accurate and timely traffic forecasting is a cornerstone of modern Intelligent Transportation Systems (ITS) because reliable short- and long-horizon predictions enable dynamic route guidance, congestion mitigation and fleet optimisation, thereby generating substantial economic and environmental benefits \cite{afandizadeh2024deep,STGCN}.  Highway and urban networks, however, produce highly nonlinear, non-stationary time series whose long-range, spatially entangled dependencies routinely defeat classical statistical models.

Recent advances in deep learning have led to the adoption of neural network architectures that model temporal dependencies. Transformer-based models have gained attention due to their ability to capture long-range dependencies using self-attention mechanisms~\cite{Vaswani2017}. Models such as Graph Multi-Attention Network (GMAN)~\cite{GMAN}, PDFormer~\cite{PDFormer}, and Spatio-Temporal AutoEncoder Transformer (STAEformer)~\cite{STAEformer} utilize multi-head attention to model spatio-temporal correlations. While effective, these models come with high computational complexity, making them less suitable for real-time forecasting applications~\cite{Kitaev2020Reformer}. Efforts like Reformer~\cite{Kitaev2020Reformer} aim to reduce this complexity, but challenges remain in the quadratic complexity of self-attention and its dependency on large training sets pose significant challenges for real-time traffic forecasting under resource constraints.

To alleviate the quadratic-cost bottleneck of Transformer forecasters, several studies have transplanted the linear-time selective state-space (Mamba)~\cite{Mamba} architecture into traffic forecasting. ST-Mamba dispenses with explicit graph convolutions and successfully recovers long-range dependencies, yet its purely implicit spatial encoding deteriorates when sensors are sparse or dynamically reconfigured~\cite{yuan2024st}. DST-Mamba augments the approach with trend–seasonal decomposition and bidirectional scanning, improving 96-step accuracy but inflating memory footprints and thus hindering edge-level deployment~\cite{he2025decomposed}. ST-MambaSync re-introduces lightweight cross-attention to sharpen spatial cues, partially forfeiting the linear scalability that originally distinguished Mamba from Transformer baselines~\cite{shao2024st}. Finally, the hierarchical resampling of ms-Mamba enhances multi-scale pattern recognition, yet its fixed down-/up-sampling ratios adapt poorly to non-stationary traffic regimes~\cite{meric2025ms}. These variants underscore the promise of selective state-space modelling, but none concurrently captures evolving temporal dynamics and adaptive graph structure under tight resource budgets.

Considering the spatial dependencies in traffic data, Graph Convolutional Networks (GCNs) have been widely adopted to model the underlying graph structure of traffic networks. GCNs~\cite{GCN} and their variants, such as Graph Attention Networks (GAT)~\cite{GAT}, have shown promise in capturing spatial correlations among traffic nodes. These models leverage the graph structure to propagate information across connected nodes, enabling them to learn complex spatial relationships. Models like Diffusion Convolutional Recurrent Neural Network (DCRNN)~\cite{DCRNN} and Graph WaveNet (GWNet)~\cite{GWNet} combine graph convolutions with temporal modeling to handle spatio-temporal data. Adaptive Graph Convolutional Recurrent Network (AGCRN)~\cite{AGCRN} further enhances adaptability by dynamically updating spatial relationships between nodes.  fail or traffic is rerouted. However, despite their success, static graph constructions and local aggregation functions lack adaptability to evolving traffic conditions, and GNNs alone cannot efficiently capture complex temporal patterns across extended time windows. Recent hybrids that graft selective state-space blocks onto graph backbones—e.g., STG-Mamba~\cite{li2024stg} and SpoT-Mamba~\cite{choi2024spot}—partially alleviate the temporal inefficiency, yet they still execute time and space reasoning in a single pass and anchor their spatial modules to fixed or random-walk graphs, leading to error spikes when sensors failures or unexpected traffic rerouting. The challenge remains to develop models that are robustly adaptive in both the spatial and temporal dimensions simultaneously.

To alleviate the twin bottlenecks of inefficient long-horizon modelling and rigid, static graph structure, we propose GAMMA-Net, a hybrid architecture that interleaves Graph Attention Networks (GAT) \cite{GAT} with the multi-axis selective state-space paradigm of Mamba \cite{Mamba}.  Specifically, (i) an initial GAT layer adaptively re-weights the physical road graph so every sensor begins with topology-aware features; (ii) a temporal Mamba block then distils long-range sequence information without incurring quadratic attention cost; (iii) a second GAT layer revisits the graph using these history-enriched representations, updating edge attentions in response to freshly uncovered dynamics; and (iv) a concluding spatial Mamba scan propagates the updated states across the network, capturing residual spatial dependencies with linear complexity.  By allowing each bout of temporal reasoning to reshape the graph—and each graph update to condition the next Mamba pass—GAMMA-Net unifies efficient long-horizon memory with fully adaptive spatial reasoning, thereby achieving robust accuracy across short, medium and long horizons while preserving the runtime and memory budget required for real-time Intelligent Transportation Systems.

To evaluate the effectiveness of GAMMA-Net, we conduct extensive experiments on benchmark traffic datasets, including METR-LA~\cite{DCRNN}, PEMS-BAY~\cite{DCRNN}, PEMS03~\cite{STSGCN}, PEMS04~\cite{STSGCN}, PEMS07~\cite{STSGCN}, and PEMS08~\cite{STSGCN}. The results demonstrate that GAMMA-Net consistently outperforms existing state-of-the-art models in terms of prediction accuracy across various time horizons. GAMMA-Net achieves up to a 16.25\% reduction in Mean Absolute Error (MAE) compared to baseline models. The ablation studies further validate the effectiveness of the individual components of GAMMA-Net, highlighting the importance of both the spatial and temporal components in achieving superior performance. 

To gain deeper insights into Mamba component, we propose a visualization method employing singular value decomposition (SVD) to deconstruct the state transition matrix of Mamba into its fundamental components. The resulting singular values not only reveal the contributions of individual components but also serve as indicators of system stability, with stability ensured when these values remain below a critical threshold. Furthermore, our analysis distinguishes between the spatial and temporal modules: the spatial module is primarily driven by a few dominant factors, whereas the temporal module captures a broader spectrum of dependencies. This SVD-based examination enhances the interpretability of the model and provides valuable insights into the design of robust frameworks for long-sequence processing.

Our contributions can be summarized as follows:

\begin{itemize} 
    \item We introduce GAMMA-Net, a novel model that effectively captures complex spatio-temporal dependencies in traffic data by integrating GAT and Mamba architectures. enhancing the model's capacity to handle dynamic traffic patterns by incorporating adaptive spatial attention mechanisms and efficient temporal modeling. 
    \item We conduct extensive experiments on benchmark traffic datasets, including METR-LA~\cite{DCRNN}, PEMS-BAY~\cite{DCRNN}, PEMS03~\cite{STSGCN}, PEMS04~\cite{STSGCN}, PEMS07~\cite{STSGCN}, and PEMS08~\cite{STSGCN}, demonstrating that GAMMA-Net outperforms existing state-of-the-art models in terms of prediction accuracy. GAMMA-Net achieves up to a 16.25\% reduction in Mean Absolute Error (MAE) compared to baseline models
    \item We propose a visualization method based on singular value decomposition (SVD) to analyze the state transition matrix of Mamba, providing insights into the contributions of individual components and system stability. This analysis enhances the interpretability of the model and informs the design of robust frameworks for long-sequence processing.
\end{itemize}

The remainder of this paper is organized as follows. In Section~\ref{sec:related_works}, we review related work in traffic forecasting and spatio-temporal modeling. Section~\ref{sec:approach} details the proposed GAMMA-Net model and its architectural components. In Section~\ref{sec:experiments}, we describe our experimental setup, present the evaluation results, and compare our model with existing approaches. Section~\ref{sec:discussion} discusses the theoretical underpinnings and practical implications of our approach, and finally, Section~\ref{sec:conclusion} concludes the paper with remarks on future research directions.

\section{Related Works}
\label{sec:related_works}

Accurate traffic forecasting has long been a challenging problem due to the complex spatio-temporal dependencies inherent in traffic data. Over the past few years, researchers have developed a variety of models to address these challenges. In this section, we review the most relevant prior work and describe how our proposed GAMMA-Net model builds upon and improves existing approaches. We organize our discussion into three main categories: Graph Neural Networks, Recurrent Architectures, and Transformer-Based Models.

Recurrent Neural Networks (RNNs) and their variants (e.g., LSTM, GRU) have been widely used to capture the temporal evolution of traffic data. For example, DCRNN~\cite{DCRNN} utilizes RNNs in combination with graph convolutions to model time-varying traffic flows. However, RNN-based approaches suffer from several well-known issues such as vanishing gradients and limited parallelism, which constrain their ability to effectively model long-term dependencies. In response to these limitations, recent studies have revisited state-space models for sequence modeling. The Selective State-Space Model (Mamba)~\cite{Mamba} is one such approach that efficiently captures long-range temporal dynamics by selectively preserving and updating state information. This method avoids the heavy computational overhead of traditional recurrent units and is particularly well suited for scenarios requiring real-time predictions.

The Transformer architecture~\cite{Vaswani2017}, with its self-attention mechanism, has recently emerged as a powerful alternative for modeling spatio-temporal dependencies in traffic forecasting. Models such as GMAN~\cite{GMAN} and PDFormer~\cite{PDFormer} leverage multi-head self-attention to capture dynamic correlations across spatial and temporal dimensions. Further, STAEformer~\cite{STAEformer} enhances the vanilla Transformer by incorporating spatio-temporal adaptive embeddings, which allow the model to adjust to complex traffic patterns. Other studies, such as STID~\cite{STID} and ST-Norm~\cite{STNorm}, have explored simplifications and normalization strategies to address the indistinguishability issues inherent in jointly modeling spatial and temporal features. Although Transformer-based methods are effective at capturing global dependencies, they generally incur high computational costs, which can impede their deployment in real-time applications.

The selective state-space architecture Mamba was originally proposed as a linear-time alternative to Transformer attention, achieving comparable accuracy on long sequences while markedly reducing memory and latency \cite{Mamba}.  Motivated by these properties, a series of traffic-oriented variants has emerged. ST-Mamba replaces explicit graph convolutions with a pure Mamba encoder and therefore captures kilometre-scale temporal dependencies, but its implicit spatial encoding deteriorates when sensors are sparse or dynamically re-wired \cite{yuan2024st}.  DST-Mamba adds trend–seasonal decomposition and bidirectional scanning to boost 96-step horizons, yet the additional reverse pass doubles memory footprints and limits edge deployment \cite{he2025decomposed}.  ST-MambaSync re-introduces lightweight cross-attention to sharpen spatial cues, partially forfeiting the linear scalability that distinguished the original model \cite{shao2024st}.  To recognise periodic patterns, ms-Mamba inserts fixed down-/up-sampling stages, but these handcrafted ratios adapt poorly to non-stationary regimes and unexpected congestion \cite{meric2025ms}.  More recent hybrids such as STG-Mamba and SpoT-Mamba graft graph convolutions onto a single Mamba pass, narrowing the temporal gap but still anchoring the spatial module to fixed or random-walk graphs, which leads to pronounced error spikes under road-network changes \cite{li2024stg,choi2024spot}.

Graph Neural Networks (GNNs) have attracted significant attention for traffic forecasting because of their ability to naturally model the non-Euclidean structure of traffic networks. Early works, such as the Diffusion Convolutional Recurrent Neural Network (DCRNN)~\cite{DCRNN}, employ diffusion convolution operations to capture spatial dependencies along with recurrent components for temporal dynamics. Graph WaveNet~\cite{GWNet} extends this idea by learning an adaptive dependency matrix through node embeddings and by incorporating dilated 1D convolutions to capture longer temporal sequences. Models like AGCRN~\cite{AGCRN} and STGCN~\cite{STGCN} further refine spatial modeling by dynamically learning node-specific parameters or by coupling graph convolutions with temporal convolutions. More recently, methods such as GTS~\cite{GTS} have proposed discrete graph structure learning techniques that infer latent relationships among multiple time series without relying on pre-defined graphs. Although these models are effective at capturing spatial structures, many still depend on static or pre-defined graph topologies, limiting their flexibility in adapting to real-time traffic dynamics.

Despite steady progress, the strands surveyed above reveal a persistent gap: no existing approach simultaneously offers (i) efficient long-horizon memory, (ii) fully adaptive graph reasoning, and (iii) a lightweight footprint suitable for real-time ITS deployments.  RNN–GNN hybrids such as DCRNN and GWNet alleviate spatial rigidity yet inherit the vanishing-gradient and parallelism limits of recurrent blocks; Transformer variants capture global context but incur prohibitive $O(L^{2})$ cost and require large training data; pure Mamba forecasters strip away quadratic attention but leave spatial cues implicit or tied to fixed adjacencies, causing accuracy to plummet under topology drift; and first-generation Mamba-GNN hybrids (e.g., STG-Mamba, SpoT-Mamba) still execute time and space reasoning in one shot, preventing newly uncovered temporal patterns from reshaping the graph and vice-versa.  Consequently, prediction error spikes are routinely observed when sensors fail, lanes close, or congestion waves propagate beyond the scale seen in training.  Closing this three-way trade-off between temporal depth, spatial adaptability and computational economy remains an open challenge—and it is precisely this deficiency that the interleaved GAMMA-Net architecture is designed to remedy.

\section{Approach}
\label{sec:approach}

\begin{figure*}[ht]
    \centering
    \includegraphics[width=\textwidth]{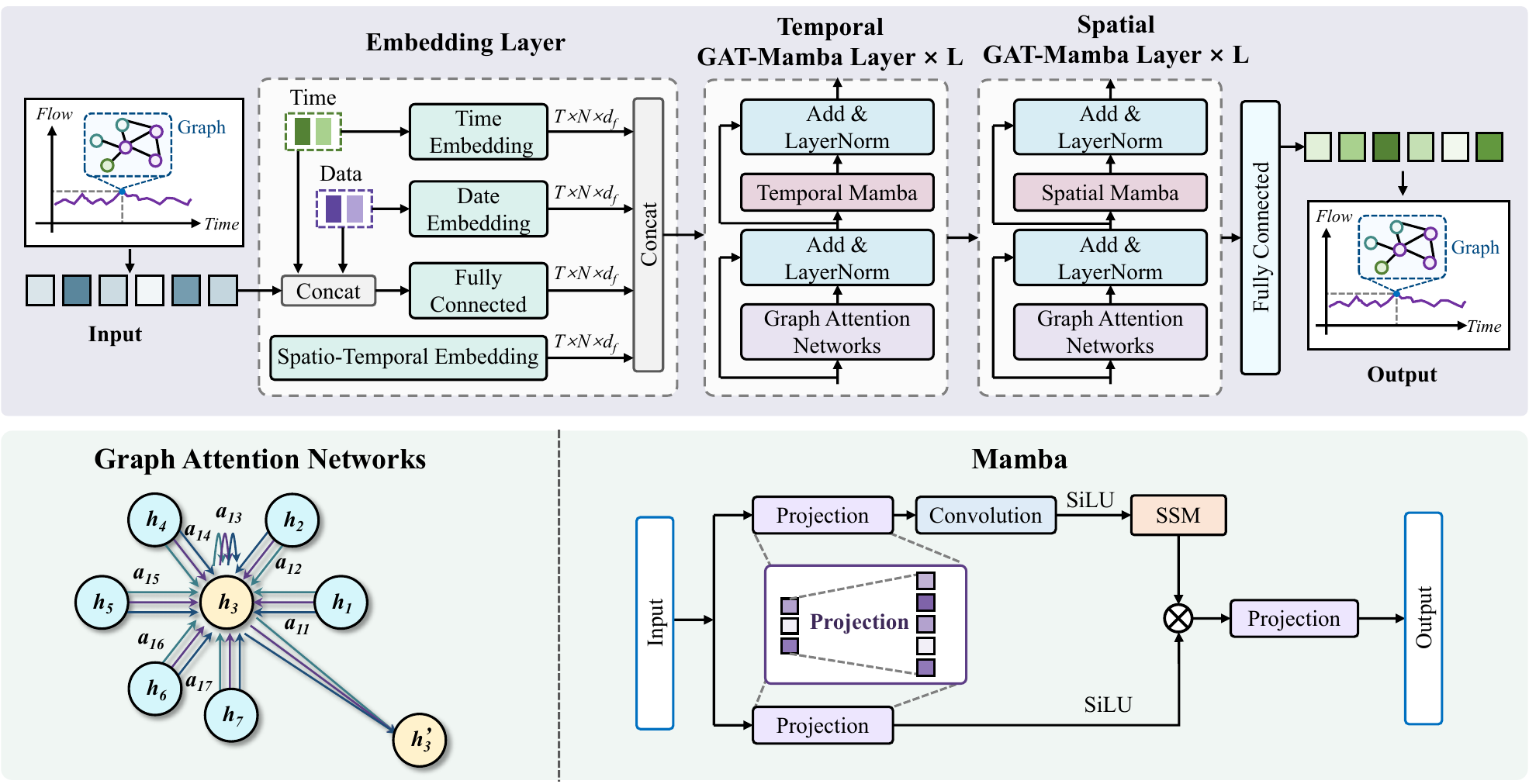}
    \caption{Model Architecture of proposed GAMMA-Net.}
    \label{image:model}
\end{figure*}

\subsection{Problem Definition}
Given the historical traffic series \(X_{t-T+1:t} \), traffic forecasting aims to predict \(T\) future graph series based on \(T^{'}\) historical series by training a model \(\mathbb{F}(\cdot)\) with parameters \(\theta\), which can be formulated as follows:  
\begin{equation}  
    [X_{t-T+1}, \dots, X_t] \xrightarrow[\theta]{\mathbb{F}(\cdot)} [X_{t+1}, \dots, X_{t+T^{'}}]  
\end{equation}  
where \(X_i \in \mathbb{R}^{N \times d}\), \(N\) denotes the number of spatial nodes and \(d\) represents the number of raw input features (flow, day, timestamp), which equals 3 in our case. Each time step \(t\) is represented by a graph \(G_t = (V, E_t, A_t)\).

\subsection{Model Architecture}
As illustrated in Figure~\ref{image:model}, GAMMA-Net comprises an embedding layer, GAMMA-Net blocks applied along the temporal axis as temporal GAMMA-Net layers and along the spatial axis as spatial GAMMA-Net layers, followed by a fully connected layer for final regression. Inspired by STAEformer~\cite{STAEformer}, the embedding layer generates a hidden representation by integrating feature embeddings, periodicity embeddings, and spatio-temporal adaptive embeddings. GAMMA-Net not only utilizes flow as the sole feature of each node but also incorporates periodicity data (time and day) as multiple features. The spatio-temporal GAMMA-Net layers capture traffic relationships, which are then processed by a regression layer to make final predictions. GAMMA-Net is purposely organised as two interleaved stacks—temporal and spatial—so that long-range sequence information is distilled *before* each update of the road-network topology and, conversely, the refreshed topology immediately conditions the next round of temporal reasoning. This section explains why each component appears where it does and how the resulting design removes the inefficiencies.

\subsubsection{Embedding Layer}
\textbf{Node Embedding:} To obtain high-dimensional representations of raw features from each node, we employ a fully connected layer to compute the embedding \(\mathbf{E}_f \in \mathbb{R}^{T \times N \times d_f}\). This fully connected layer is applied independently to each node at every time step, as defined in Equation~\ref{eq:feature_embedding}:

\begin{equation}
    \mathbf{E}_f^{(i,t)} = \mathbf{W} \mathbf{X}_t^{(i)} + \mathbf{b},
    \label{eq:feature_embedding}
\end{equation}
where \(\mathbf{X}_t^{(i)} = \begin{bmatrix} flow_t^i & time_t & day_t \end{bmatrix}^\top \in \mathbb{R}^{3}\) represents the input raw features for node \(i\) at time \(t\). \(\mathbf{W} \in \mathbb{R}^{d_f \times 3}\) and \(\mathbf{b} \in \mathbb{R}^{d_f}\) are the learnable weight matrix and bias vector, respectively.

\textbf{Periodicity Embedding:} To retain the positional information of the time series, we utilize learnable embedding matrices for both the day-of-week and timestamp-of-day features, denoted as \( \mathbf{T}_w \in \mathbb{R}^{N_w \times d_f} \) and \( \mathbf{T}_d \in \mathbb{R}^{N_d \times d_f} \), respectively. Specifically, for each time step, the corresponding day-of-week and timestamp-of-day embeddings are retrieved from \( \mathbf{T}_w \) and \( \mathbf{T}_d \). These embeddings are then concatenated and broadcasted to form the periodicity embedding \( \mathbf{E}_p \in \mathbb{R}^{T \times N \times 2d_f} \) for the traffic time series. This periodicity embedding \( \mathbf{E}_p \) effectively encodes both weekly and daily temporal patterns, enhancing the ability of model to capture recurring trends in the traffic data.

\textbf{Spatiotemporal Adaptive Embedding:} Temporal relationships in traffic time series are governed not only by periodicity but also by chronological order~\cite{STAEformer}. For instance, a time frame within the traffic time series is more similar to adjacent time frames, whereas time series from different sensors typically exhibit distinct temporal patterns. Therefore, we adopt a spatiotemporal adaptive embedding, first introduced in STAEformer~\cite{STAEformer}, \( \mathbf{E}_a \in \mathbb{R}^{T \times N \times d_a} \), to capture intricate spatiotemporal relationships.

By concatenating the embeddings above, we obtain the hidden spatio-temporal representation \( Z \in \mathbb{R}^{T \times N \times d_h} \) where the hidden dimension \( d_h =  3d_f + d_a \).

\subsubsection{Graph Construction and Optimization}
\label{sec:graph_construction}
In our framework, each node in the graph corresponds to a traffic sensor or a monitored road segment and is associated with features such as real-time traffic flow, timestamps, and periodicity information. These features enable the model to capture both instantaneous traffic conditions and recurring temporal patterns. Edges are constructed exclusively on the basis of direct physical connections present in the actual traffic network. In other words, an edge is established between two nodes if and only if they are directly connected in the traffic network—for instance, when they are located on the same road or are directly linked at an intersection. This method of edge construction ensures that the graph faithfully represents the true topology of the traffic system, thereby facilitating accurate propagation and fusion of traffic information. The static, predefined graph structure leverages domain-specific knowledge to provide a solid structural basis.

\subsubsection{GAT-Mamba Pair}
The GAT-Mamba pair constitutes the central processing component of the proposed GAMMA-Net. It is specifically designed to address the limitations of prior spatio-temporal traffic forecasting models—particularly the difficulty of capturing long-range temporal dependencies and adaptively modeling dynamic spatial relationships under computational constraints. Existing approaches often rely on fixed graph structures that cannot accommodate evolving traffic dynamics or incorporate computationally intensive mechanisms such as Transformers. Moreover, earlier hybrid models combining graph neural networks with Mamba architectures frequently process temporal and spatial information in a sequential, unidirectional manner without feedback, resulting in static graph structures that are not revised based on newly inferred temporal patterns.

To overcome these deficiencies, the GAMMA-Net block introduces an interleaved architecture that integrates Graph Attention Networks (GAT) \cite{GAT} and selective state-space model scans (Mamba) \cite{Mamba} in a structured sequence:
\begin{equation}
(\text{GAT} \rightarrow \text{Mamba}*{\text{Temporal}})*{L} \rightarrow (\text{GAT} \rightarrow \text{Mamba}*{\text{Spatial}})*{L}
\end{equation}
where $L$ denotes the number of stacked GAT-Mamba pairs.

\textbf{Stage 1: Initial Graph Attention followed by Temporal State-Space Scan.}
Given hidden representations $Z \in \mathbb{R}^{T \times N \times d_h}$, a multi-head GAT layer is first applied at each time step $t$ to adaptively modulate the graph structure based on node features:
\begin{equation}
H^{(t)} = \operatorname{GAT}(G, Z^{(t)}), \quad t = 1, \dots, T
\end{equation}
This step dynamically re-weights the graph $G(V,E)$, allowing the model to emphasize contextually relevant spatial dependencies. Unlike models based on fixed graph convolutions, GAT enables the graph structure to respond to current traffic conditions, producing topology-aware, context-sensitive embeddings that are crucial for subsequent temporal modeling. Residual connection and layer normalization are then applied to yield the updated representation:
\begin{equation}
\hat{H} = \operatorname{LayerNorm}(H + Z)
\end{equation}

To capture long-range temporal dependencies with linear computational complexity, the spatially enriched features $\hat{H}$ are processed using a Mamba scan along the temporal dimension:
\begin{equation}
M = \operatorname{Mamba}\_{\text{Temporal}}(\hat{H})
\end{equation}
This approach mitigates the quadratic complexity of Transformer-based models and surpasses the memory limitations of recurrent architectures, resulting in an efficient representation of temporal trends. The final temporally contextualized representation is obtained via a residual connection and layer normalization, denoted as $Z_{\text{time}}$.

\textbf{Stage 2: Second Graph Attention followed by Spatial State-Space Scan.}
To further refine the spatial representation in light of the temporal context—such as detecting emerging congestion patterns—$Z_{\text{time}}$ is fed into a second GAT layer:
\begin{equation}
S = \operatorname{GAT}(G, Z\_{\text{time}})
\end{equation}
This step enables dynamic recalibration of the graph attention weights based on temporal insights, overcoming the common shortcoming of static graph structures in previous methods.

Finally, to efficiently propagate the temporally-informed spatial information across the network topology, a Mamba scan is applied along the spatial dimension:
\begin{equation}
Z' = \operatorname{Mamba}\_{\text{Spatial}}(S)
\end{equation}
This operation enables efficient spatial mixing without resorting to computationally expensive graph convolutions. Leveraging Mamba’s linear complexity, this step ensures scalability while capturing refined spatial dependencies aligned with the updated attention-modulated graph.

\subsubsection{Regression Layer}
To generate the final predictions, we reshape the output of the spatio-temporal GAMMA-Net layers \(\mathbf{Z}' \in \mathbb{R}^{T \times N \times d_h}\) by flattening the temporal and feature dimensions for each node, resulting in \(\mathbf{Z}'' \in \mathbb{R}^{N \times (T \times d_h)}\). The regression layer then computes the predictions as:

\begin{equation}
\hat{\mathbf{Y}} = \mathbf{Z}'' \mathbf{W}^\top + \mathbf{b},
\end{equation}
where \(\hat{\mathbf{Y}} \in \mathbb{R}^{N \times T'}\) is the predicted output for each node over the prediction horizon \(T'\), \(\mathbf{W} \in \mathbb{R}^{T' \times (T \times d_h)}\) represents the learnable weights, and \(\mathbf{b} \in \mathbb{R}^{T'}\) is the bias vector.

\section{Experiments}
\label{sec:experiments}
\begin{figure*}[ht]
    \centering
    \subfigure[METR-LA] {\includegraphics[height=0.35\textwidth]{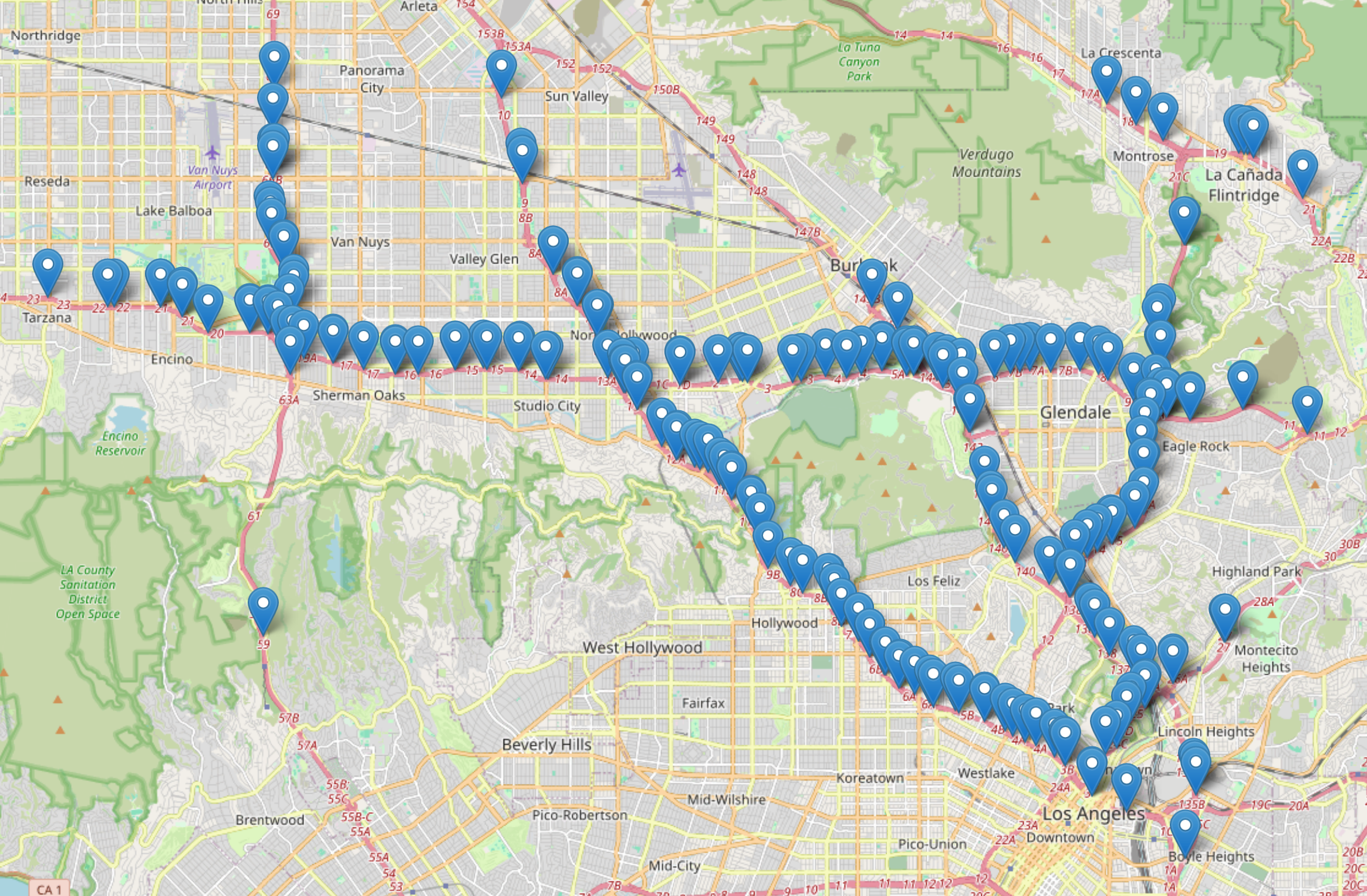}}
    \quad \quad
    \subfigure[PEMS-BAY] {\includegraphics[height=0.35\textwidth]{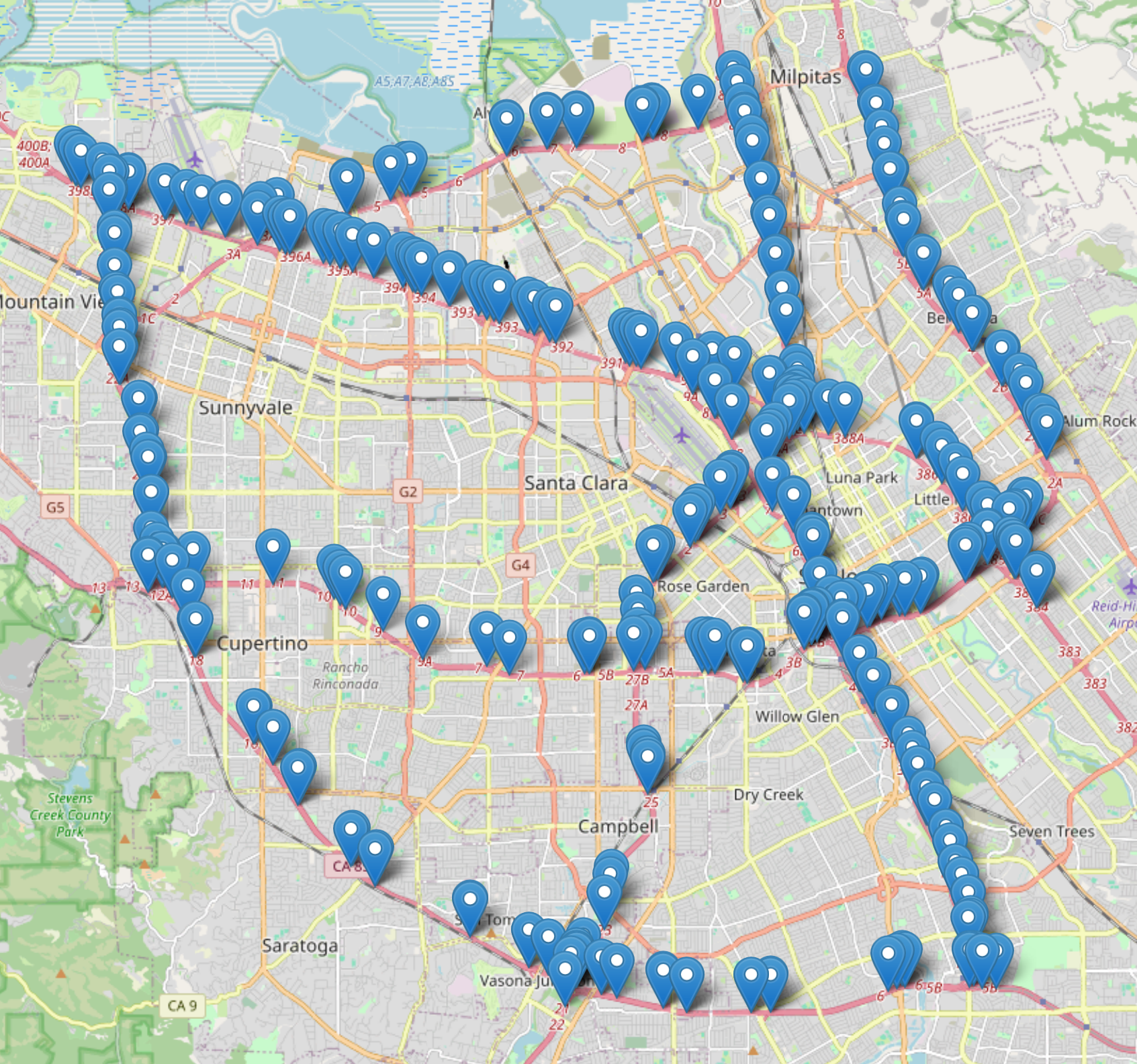}}
    \caption{Node distribution of the METR-LA (a) and PEMS-BAY dataset (b).}
    \label{image:dataset}
\end{figure*}

\subsection{Benchmark datasets}
To assess the performance of GAMMA-Net, we employ several benchmark traffic datasets, including METR-LA~\cite{DCRNN}, PEMS-BAY~\cite{DCRNN}, PEMS03~\cite{STSGCN}, PEMS04~\cite{STSGCN}, PEMS07~\cite{STSGCN}, and PEMS08~\cite{STSGCN}. These datasets are commonly used in traffic forecasting and analysis, especially within the context of graph neural networks and time-series modeling. Table~\ref{tab:datasets} provides a summary of these datasets, and Figure~\ref{image:dataset} further visualizes the node distribution of the METR-LA and PEMS-BAY dataset.

\begin{table*}[htbp]
    \centering
    \caption{Summary of Traffic Datasets}
    \label{tab:datasets}
    \resizebox{\textwidth}{!}{
        \begin{tabular}{ccccc}
        \toprule
        \textbf{Dataset} & \textbf{Sampling Interval} & \textbf{Timesteps} & \textbf{Nodes (N)} & \textbf{Region} \\
        \midrule
        METR-LA    & 5 minutes & 34,272 & 207 & Los Angeles County highways \\
        PEMS-BAY   & 5 minutes & 52,116 & 325 & San Francisco Bay Area freeways \\
        PEMS03     & 5 minutes & 26,209 & 358 & California District 3 \\
        PEMS04     & 5 minutes & 16,992 & 307 & California District 4 \\
        PEMS07     & 5 minutes & 28,224 & 883 & California District 7 \\
        PEMS08     & 5 minutes & 17,856 & 170 & California District 8 \\
        \bottomrule
        \end{tabular}
    }
\end{table*}

\subsection{Implementation}
The proposed model, GAMMA-Net, was implemented using the open-source software library PyTorch (version 2.1.1) on a workstation equipped with a single NVIDIA GeForce RTX 4090 GPU. METR-LA and PEMS-BAY datasets are divided into training, validation, and test sets in a ratio of 7:1:2. PEMS03, PEMS04, PEMS07, and PEMS08 are divided into these sets in a ratio of 6:2:2. Specifically, the embedding dimension \(d_f\) is set to 24, and \(d_a\) is set to 80. The number of layers \(num\_layers\) is 3 for both the spatial and temporal GAMMA-Net layers. The number of attention heads \(num\_heads\) in GAT is set to 4. The Adam optimizer is used with an initial learning rate of \(1 \times 10^{-3}\), and the batch size is set to 16. We apply an early-stopping mechanism when the validation error converges within 30 consecutive steps. The best epochs for METR-LA, PEMS-BAY, PEMS03, PEMS04, PEMS07, and PEMS08 are 21, 12, 27, 39, 37, and 47, respectively.

\subsection{Metrics}
We use three widely used metrics for the traffic forecasting task: Mean Absolute Error (MAE, Equation~\ref{mae}), Root Mean Squared Error (RMSE, Equation~\ref{rmse}), and Mean Absolute Percentage Error (MAPE, Equation~\ref{mape}). Following previous work, we report the average performance over all 12 predicted horizons on the PEMS03, PEMS04, PEMS07, and PEMS08 datasets. For the METR-LA and PEMS-BAY datasets, we evaluate performance at horizons 3, 6, and 12 (corresponding to 15, 30, and 60 minutes).

\begin{equation}
    \label{mae}
    \mathrm{MAE} = \frac{1}{n} \sum_{i=1}^{n} \left| y_i - \hat{y}_i \right| 
\end{equation}

\begin{equation} 
    \label{rmse} 
    \mathrm{RMSE} = \sqrt{ \frac{1}{n} \sum_{i=1}^{n} \left( y_i - \hat{y}_i \right)^2 } 
\end{equation}

\begin{equation}
    \label{mape} 
    \mathrm{MAPE} = \frac{1}{n} \sum_{i=1}^{n} \left| \frac{ y_i - \hat{y}_i }{ y_i } \right| 
\end{equation}

\subsection{Performance Evaluation}
\begin{table*}[ht]
    \centering
    \caption{Performance of various models on METR-LA and PEMS-BAY datasets.}
    \label{tab:metrics1}
    \resizebox{\textwidth}{!}{
        \begin{tabular}{lcccccccccccccc}
            \toprule
            & \multirow{2}{*}{\textbf{Model}}& \multicolumn{3}{c}{\textbf{Horizon 3 (15 min)}} &~ & \multicolumn{3}{c}{\textbf{Horizon 6 (30 min)}} & ~ &\multicolumn{3}{c}{\textbf{Horizon 12 (60 min)}} \\
            \cmidrule{3-5}\cmidrule{7-9}\cmidrule{11-13}
            ~ & ~ & \textbf{MAE} & \textbf{RMSE} & \textbf{MAPE} & ~ & \textbf{MAE} & \textbf{RMSE} & \textbf{MAPE} &~ & \textbf{MAE} & \textbf{RMSE} & \textbf{MAPE} \\
            \midrule
            \multirow{13}{*}{\rotatebox{90}{METR-LA}}
            & GWNet~\cite{GWNet}  & 2.69 & 5.15 & 6.99\% &~ & 3.08 & 6.20 & 8.47\% & ~ &3.51 & 7.28 & 9.96\% \\
            & DCRNN~\cite{DCRNN} & 2.67 & 5.16 & 6.86\% &~ & 3.12 & 6.27 & 8.42\% &~ & 3.54 & 7.47 & 10.32\% \\
            & AGCRN~\cite{AGCRN}& 2.85 & 5.53 & 7.63\% &~ & 3.20 & 6.52 & 9.00\% & ~ &3.59 & 7.45 & 10.47\% \\
            & STGCN~\cite{STGCN}& 2.75 & 5.29 & 7.12\% & ~ &3.15 & 6.35 & 8.62\% & ~ &3.60 & 7.43 & 10.35\% \\
            & GTS~\cite{GTS} & 2.75 & 5.27 & 6.89\% &~ & 3.14 & 6.33 & 8.16\% &~ & 3.59 & 7.44 & 10.25\% \\
            & MTGNN~\cite{MTGNN}& 2.69 & 5.16 & 7.40\% &~ & 3.05 & 6.13 & 8.47\% & ~ &3.47 & 7.21 & 9.70\% \\
            & STNorm ~\cite{STNorm}& 2.81 & 5.57 & 7.41\% & ~ &3.18 & 6.59 & 10.24\% & ~ &3.57 & 7.51 & 10.24\% \\
            & GMAN~\cite{GMAN}& 2.80 & 5.55 & 7.77\% &~ & 3.12 & 6.49 & 8.73\% & ~ &3.44 & 7.35 & 10.07\% \\
            & PDFormer~\cite{PDFormer}& 2.83 & 5.45 & 7.75\% &~ & 3.20 & 6.46 & 9.19\% &~ & 3.62 & 7.47 & 10.91\% \\
            & STID~\cite{STID}& 2.82 & 5.53 & 6.85\% &~ & 3.19 & 6.57 & 9.39\% & ~ & 3.55 & 7.55 & 10.95\% \\
            & STAEformer~\cite{STAEformer} & 2.65 & 5.11 & 6.85\% & ~ &2.97 & 6.00 & 8.13\% &~ & 3.34 & \textcolor{blue}{\textbf{7.02}} & 9.70\% \\
            & STGM~\cite{lablack2023spatio} &  \textcolor{blue}{\textbf{2.57}} & \textcolor{blue}{\textbf{4.89}} & \textcolor{blue}{\textbf{6.52\%}} & ~ &\textcolor{blue}{\textbf{2.86}} & \textcolor{blue}{\textbf{5.76}} & \textcolor{blue}{\textbf{7.80\%}} &~ & \textcolor{blue}{\textbf{3.23}} & 7.10 & \textcolor{blue}{\textbf{9.39\%}} \\
            & ST-MambaSync~\cite{shao2024st} & 2.63 & 5.05 & 6.80\% & ~ &2.91 & 6.07 & 8.08\% &~ & 3.31 & \textcolor{blue}{\textbf{7.02}} & 9.70\% \\
            & \textbf{GAMMA-Net} & \textcolor{red}{\textbf{2.39}}  & \textcolor{red}{\textbf{4.54}}  & \textcolor{red}{\textbf{6.22\%}} &~ & \textcolor{red}{\textbf{2.60}} & \textcolor{red}{\textbf{5.20}} & \textcolor{red}{\textbf{7.06\%}} &~& \textcolor{red}{\textbf{2.87}} & \textcolor{red}{\textbf{5.99}} & \textcolor{red}{\textbf{8.16\%}} \\
    
            \midrule
            \multirow{13}{*}{\rotatebox{90}{PEMS-BAY}}
            & GWNet~\cite{GWNet}& 1.30 & 2.73 & 2.71\% & ~ &1.63 & 3.73 & 3.73\% &~ & 1.99 & 4.60 & 4.71\% \\
            & DCRNN~\cite{DCRNN}& 1.31 & 2.76 & 2.73\% &~ & 1.65 & 3.75 & 3.71\% & ~ &1.97 & 4.60 & 4.68\% \\
            & AGCRN~\cite{AGCRN}& 1.35 & 2.88 & 2.91\% & ~ &1.67 & 3.82 & 3.81\% & ~ &1.94 & 4.50 & 4.55\% \\
            & STGCN~\cite{STGCN}& 1.36 & 2.88 & 2.86\% & ~ &1.70 & 3.84 & 3.79\% & ~ &2.02 & 4.63 & 4.72\% \\
            & GTS~\cite{GTS}& 1.37 & 2.92 & 2.85\% &~ & 1.72 & 3.86 & 3.88\% &~ & 2.06 & 4.60 & 4.88\% \\
            & MTGNN~\cite{MTGNN}& 1.33 & 2.80 & 2.81\% & ~ &1.66 & 3.77 & 3.75\% & ~ &1.95 & 4.50 & 4.62\% \\
            & STNorm~\cite{STNorm}& 1.33 & 2.82 & 2.76\% &~ & 1.65 & 3.77 & 3.66\% & ~ &1.92 & 4.45 & 4.46\% \\
            & GMAN~\cite{GMAN}& 1.35 & 2.90 & 2.87\% & ~ &1.65 & 3.82 & 3.74\% & ~ &1.92 & 4.49 & 4.52\% \\
            & PDFormer~\cite{PDFormer}& 1.32 & 2.83 & 2.78\% &~ & 1.64 & 3.79 & 3.71\% & ~ &1.91 & 4.43 & 4.51\% \\
            & STID~\cite{STID}& 1.31 & 2.79 & 2.78\% &~ & 1.64 & 3.73 & 3.73\% &~ & 1.91 & 4.42 & 4.55\% \\
            & STAEformer~\cite{STAEformer} & 1.31 & 2.78 & 2.76\% & ~ &1.62 & 3.68 & 3.62\% &~ & 1.88 & 4.34 & 4.41\% \\
            & STGM~\cite{lablack2023spatio} &  \textcolor{blue}{\textbf{1.25}} & \textcolor{blue}{\textbf{2.62}} & \textcolor{blue}{\textbf{2.70\%}} & ~ &\textcolor{blue}{\textbf{1.60}} & 3.69 & \textcolor{blue}{\textbf{3.55\%}} &~ & \textcolor{blue}{\textbf{1.86}} & 4.37 & \textcolor{blue}{\textbf{4.34\%}} \\
            & ST-MambaSync~\cite{shao2024st} & 1.30 & 2.75 & 2.75\% & ~ &1.63 & \textcolor{blue}{\textbf{3.62}} & 3.61\% &~ & 1.87 & \textcolor{blue}{\textbf{4.30}} & 4.40\% \\
            & \textbf{GAMMA-Net} & \textcolor{red}{\textbf{1.12}} &	\textcolor{red}{\textbf{2.28}} 	&\textcolor{red}{\textbf{2.31\%}} &	~ &\textcolor{red}{\textbf{1.34}}	&\textcolor{red}{\textbf{2.97}} 	&\textcolor{red}{\textbf{2.91\%}} &~ 	&\textcolor{red}{\textbf{1.59}} 	&\textcolor{red}{\textbf{3.67}} &\textcolor{red}{\textbf{3.61\%}} \\
            \bottomrule
        \end{tabular}
    }
    \footnotesize{The best results are highlighted in \textcolor{red}{\textbf{red}} and the second-best results are highlighted in \textcolor{blue}{\textbf{blue}}.}
\end{table*}

To evaluate the effectiveness of the proposed model, GAMMA-Net, we select several state-of-the-art spatio-temporal forecasting models as baselines, including graph neural networks (GNNs), recurrent neural networks (RNNs), and Transformer-based models. These models have demonstrated excellent performance in forecasting spatio-temporal data, particularly in applications like traffic flow. 

Table~\ref{tab:metrics1} presents the performance of GAMMA-Net and baseline models on the METR-LA and PEMS-BAY datasets across different prediction horizons, and Table~\ref{tab:metrics2} provides the results for the PEMS03, PEMS04, PEMS07, and PEMS08 datasets. The results show that GAMMA-Net consistently outperforms existing models in terms of MAE, RMSE, and MAPE, achieving up to a 16.25\% reduction in MAE compared to baseline models.

\begin{figure*}[ht]
    \centering
    \includegraphics[width=0.9\textwidth]{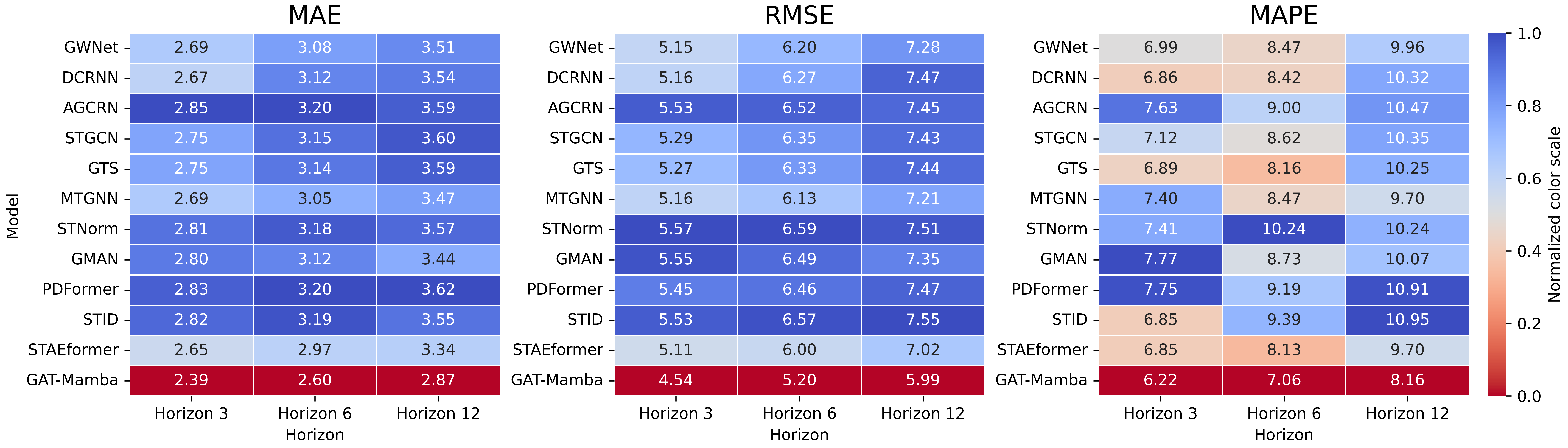}
    \caption{Performance heatmaps of twelve predictive models on the Metr-LA dataset for forecasting horizons of 3, 6, and 12 time steps, showing MAE (left), RMSE (center) and MAPE (right).  Each row corresponds to a model and each column to a forecast horizon.  Cell shading reflects Min–Max normalization of the respective metric (darker colours denote lower error), while the overlaid annotations give the original error values.}
    \label{image:mertla-metrics}
\end{figure*}

\begin{figure*}[ht]
    \centering
    \includegraphics[width=0.9\textwidth]{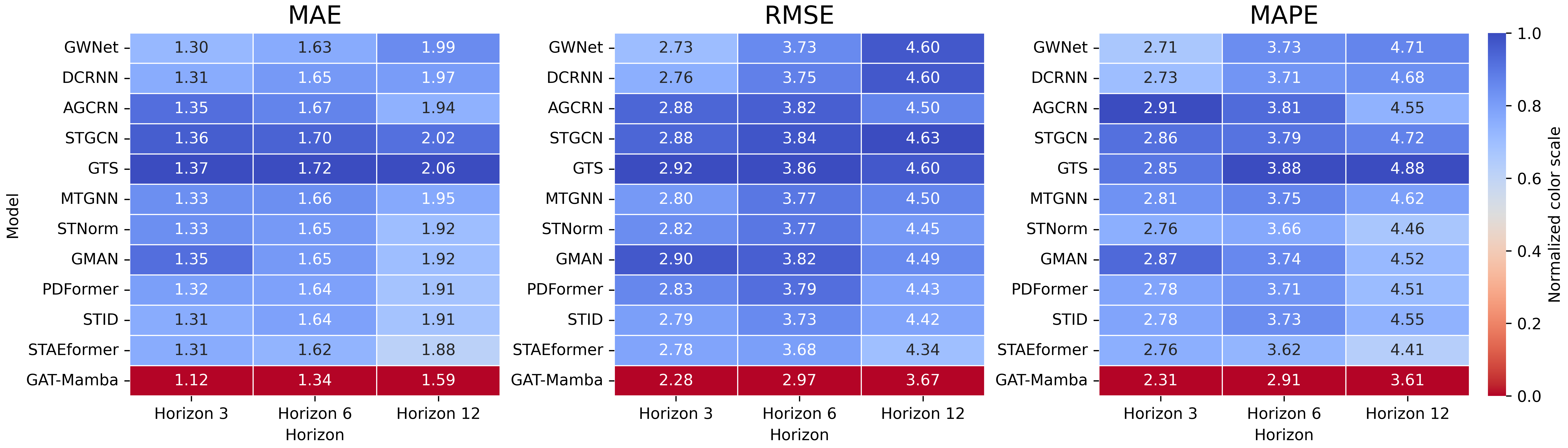}
    \caption{Performance heatmaps of twelve predictive models on the PEMS-Bay dataset for forecasting horizons of 3, 6, and 12 time steps, showing MAE (left), RMSE (center) and MAPE (right).  Each row corresponds to a model and each column to a forecast horizon.  Cell shading reflects Min–Max normalization of the respective metric (darker colours denote lower error), while the overlaid annotations give the original error values.}
    \label{image:pemsbay-metrics}
\end{figure*}

\begin{figure*}[ht]
    \centering
    \includegraphics[width=0.9\textwidth]{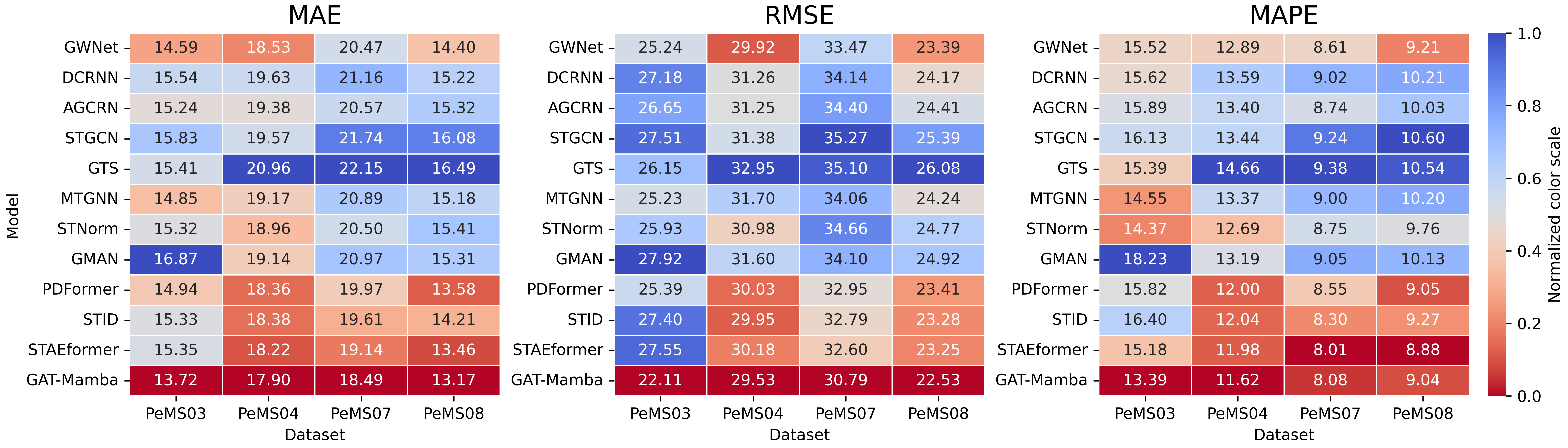}
    \caption{Performance heatmaps of twelve predictive models across four traffic sub-datasets (PeMS03, PeMS04, PeMS07, and PeMS08), showing MAE (left), RMSE (center) and MAPE (right).  Each row corresponds to a model and each column to a sub-dataset.  Cell shading reflects Min–Max normalization of the respective metric (darker colours denote lower error), while the overlaid annotations give the original error values.}
    \label{image:pems-metrics}
\end{figure*}

Figure~\ref{image:mertla-metrics} and Figure~\ref{image:pemsbay-metrics} illustrate the performance of GAMMA-Net and baseline models on the METR-LA and PEMS-BAY datasets, respectively. The figures show that GAMMA-Net consistently outperforms existing models across all prediction horizons, further validating its effectiveness in capturing complex spatio-temporal dependencies. Figure~\ref{image:pems-metrics} shows the performance of GAMMA-Net and baseline models on the PEMS03, PEMS04, PEMS07, and PEMS08 datasets. The results further indicate that GAMMA-Net achieves significant improvements in MAE, RMSE, and MAPE compared to existing models.

GAMMA-Net model sets a new benchmark in traffic forecasting, outperforming established models across multiple horizons and datasets. Its innovative use of graph attention and selective state spaces provides more accurate, stable, and reliable predictions, making it a valuable tool for advanced traffic management and planning. The model's scalability and computational efficiency further underscore its applicability in real-world scenarios, promising substantial improvements in traffic system operations.

\begin{table*}[htbp]
    \centering
    \caption{Performance on PEMS03, 04, 07, and 08.}
    \label{tab:metrics2}
    \resizebox{\textwidth}{!}{
        \begin{tabular}{cccccccccccccccc}
            \toprule
        \multirow{2}{*}{\textbf{Model}} & \multicolumn{3}{c}{\textbf{PEMS03}} &~ & \multicolumn{3}{c}{\textbf{PEMS04}} & ~ &\multicolumn{3}{c}{\textbf{PEMS07}} &~ & \multicolumn{3}{c}{\textbf{PEMS08}} \\ 
        \cmidrule{2-4}\cmidrule{6-8}\cmidrule{10-12}\cmidrule{14-16}
    
        ~  & \textbf{MAE} & \textbf{RMSE} & \textbf{MAPE} &~ & \textbf{MAE} & \textbf{RMSE} & \textbf{MAPE} & ~ &\textbf{MAE} & \textbf{MSE} & \textbf{MAPE} &~ & \textbf{MAE} & \textbf{RMSE} & \textbf{MAPE} \\
        \midrule
    
        GWNet~\cite{GWNet}            & 14.59 & 25.24 & 15.52\% &~ & 18.53 & 29.92 & 12.89\% &~ & 20.47 & 33.47 & 8.61\%  &~ & 14.40 & 23.39 & 9.21\%  \\
        DCRNN~\cite{DCRNN}           & 15.54 & 27.18 & 15.62\% &~ & 19.63 & 31.26 & 13.59\% &~ & 21.16 & 34.14 & 9.02\%  &~ & 15.22 & 24.17 & 10.21\% \\
        AGCRN~\cite{AGCRN}         & 15.24 & 26.65 & 15.89\% &~ & 19.38 & 31.25 & 13.40\% &~ & 20.57 & 34.40 & 8.74\%  &~ & 15.32 & 24.41 & 10.03\% \\
        STGCN~\cite{STGCN}           & 15.83 & 27.51 & 16.13\% &~ & 19.57 & 31.38 & 13.44\% &~ & 21.74 & 35.27 & 9.24\%  &~ & 16.08 & 25.39 & 10.60\% \\
        GTS~\cite{GTS}             & 15.41 & 26.15 & 15.39\% &~ & 20.96 & 32.95 & 14.66\% &~ & 22.15 & 35.10 & 9.38\%  &~ & 16.49 & 26.08 & 10.54\% \\
        MTGNN~\cite{MTGNN}           & 14.85 & 25.23 & 14.55\% &~ & 19.17 & 31.70 & 13.37\% &~ & 20.89 & 34.06 & 9.00\%  &~ & 15.18 & 24.24 & 10.20\% \\
        STNorm~\cite{STNorm}          & 15.32 & 25.93 & 14.37\% &~ & 18.96 & 30.98 & 12.69\% &~ & 20.50 & 34.66 & 8.75\%  &~ & 15.41 & 24.77 & 9.76\%  \\
        GMAN~\cite{GMAN}            & 16.87 & 27.92 & 18.23\% &~ & 19.14 & 31.60 & 13.19\% &~ & 20.97 & 34.10 & 9.05\%  &~ & 15.31 & 24.92 & 10.13\% \\
        PDFormer~\cite{PDFormer}        & 14.94 & 25.39 & 15.82\% &~ & 18.36 & 30.03 & 12.00\% &~ & 19.97 & 32.95 & 8.55\%  &~ & 13.58 & 23.41 & 9.05\%  \\
        STID~\cite{STID}            & 15.33 & 27.40 & 16.40\% &~ & 18.38 & 29.95 & 12.04\% &~ & 19.61 & 32.79 & 8.30\%  &~ & 14.21 & 23.28 & 9.27\%  \\
        STAEformer~\cite{STAEformer}   & 15.35 &  27.55 & 15.18\% &~ & 18.22 & 30.18 & 11.98\% &~ & 19.14 & 32.60 & 8.01\%  &~ & 13.46 & 23.25 & 8.88\%  \\
        STD-MAE~\cite{gao2023spatial} & \textcolor{blue}{\textbf{13.80}} & \textcolor{blue}{\textbf{24.43}} & \textcolor{blue}{\textbf{13.96\%}} &~ &\textcolor{red}{\textbf{17.80}} & \textcolor{red}{\textbf{29.25}} & \textcolor{blue}{\textbf{11.97\%}} &~ & \textcolor{blue}{\textbf{18.65}} & \textcolor{blue}{\textbf{31.44}} & \textcolor{red}{\textbf{7.84\%}} &~& 13.44 & \textcolor{red}{\textbf{22.47}} & \textcolor{red}{\textbf{8.76\%}} \\
        ST-MambaSync~\cite{shao2024st} & 15.30 &27.47 &15.18\%&~ &18.20& 29.85 &12.00\%&~& 19.14 &32.58 &\textcolor{blue}{\textbf{7.97\%}}&~ &\textcolor{blue}{\textbf{13.30}} &23.14 &\textcolor{blue}{\textbf{8.80\%}} \\
        \textbf{GAMMA-Net}       & \textcolor{red}{\textbf{13.72}} & \textcolor{red}{\textbf{22.11}} & \textcolor{red}{\textbf{13.39\%}} &~ & \textcolor{blue}{\textbf{17.90}}   & \textcolor{blue}{\textbf{29.53}}  & \textcolor{red}{\textbf{11.62\%}} &~ &\textcolor{red}{\textbf{18.49}}  & \textcolor{red}{\textbf{30.79}}  & 8.08\%  &~ & \textcolor{red}{\textbf{13.17}}  & \textcolor{blue}{\textbf{22.53}}  & 9.04\% \\
        \bottomrule
        \end{tabular}
    }
    \footnotesize{The best results are highlighted in \textcolor{red}{\textbf{red}} and the second-best results are highlighted in \textcolor{blue}{\textbf{blue}}.}
\end{table*}
    
\subsection{Ablation Study}
An ablation study is conducted on the GAMMA-Net model to evaluate the importance of its individual components by analyzing their contributions to performance on the METR-LA and PEMS-BAY datasets, as detailed in Table~\ref{tab:ablation}.

\begin{table*}[htbp]
    \centering
    \caption{Ablation Study on METR-LA and PEMS-BAY datasets}
    \label{tab:ablation}
    \resizebox{\textwidth}{!}{
        \begin{tabular}{lcccccccccccccc}
            \toprule
            & \multirow{2}{*}{\textbf{Model}}& \multicolumn{3}{c}{\textbf{Horizon 3 (15 min)}} &~ & \multicolumn{3}{c}{\textbf{Horizon 6 (30 min)}} & ~ &\multicolumn{3}{c}{\textbf{Horizon 12 (60 min)}} \\
            \cmidrule{3-5}\cmidrule{7-9}\cmidrule{11-13}
            ~ & ~ & \textbf{MAE} & \textbf{RMSE} & \textbf{MAPE} & ~ & \textbf{MAE} & \textbf{RMSE} & \textbf{MAPE} &~ & \textbf{MAE} & \textbf{RMSE} & \textbf{MAPE} \\
            \midrule
            \multirow{5}{*}{\rotatebox{90}{METR-LA}}
            & w/o GAT & 2.42 &	4.64 &	6.29\% &~ &	2.65 &	5.35 &	7.24\% &	~ &2.94 &	6.19 &	8.41\%  \\
            & w/o Temporal GAMMA-Net & 2.42 &	4.61 &	6.36\% &	~ &2.63 &	5.26 &	7.24\% &~ &	2.90 &	6.05 &	8.34\%  \\
            & w/o Spatio GAMMA-Net & 2.41 	&4.59 &	6.22\% &	~ &2.61 &	5.22 &	7.03\% &	~ &2.88 &	5.99 &	8.11\%  \\
            & w/o both GAMMA-Net & 3.03 &	6.36 &	7.46\% &~ &	3.47 &	7.44 &	8.99\% &~ &	4.14 &	8.62 &	11.68\%  \\
            & \textbf{GAMMA-Net} & \textbf{2.39}  & \textbf{4.54}  & \textbf{6.22\%} &~ & \textbf{2.60} & \textbf{5.20} & \textbf{7.06\%} &~& \textbf{2.87} & \textbf{5.99} & \textbf{8.16\%} \\
    
            \midrule
            \multirow{5}{*}{\rotatebox{90}{PEMS-BAY}}
            & w/o GAT & 1.14 &	2.32 &	2.39\% &	~ &	1.36 &	3.02 &	3.01\% &	~ &	1.60 	&3.71 	&3.71\%  \\
            & w/o Temporal GAMMA-Net & 1.13 &	2.30 	&2.35\% &	~ &	1.35 	&2.98 &	2.95\% &	~ 	&1.60 	&3.68 &	3.63\%  \\
            & w/o Spatio GAMMA-Net & 1.16 &	2.41 &	2.43\% &	~& 	1.38 &	3.06 &	3.02\% &	~ 	&1.61 &	3.70 	&3.68\%  \\
            & w/o both GAMMA-Net & 1.35 &	2.86 &	2.77\% &	~ &	1.73 &	3.92 &	3.74\% &	~ &	2.31 &	5.35 &	5.39\%  \\
            & \textbf{GAMMA-Net} & \textbf{1.12} &	\textbf{2.28} 	&\textbf{2.31\%} &	~ &\textbf{1.34}	&\textbf{2.97} 	&\textbf{2.91\%} &~	&\textbf{1.59} 	&\textbf{3.67} &\textbf{3.61\%} \\
            \bottomrule
        \end{tabular}
    }
\end{table*}

\begin{itemize}
    \item \textbf{GAMMA-Net without Graph Attention Components (w/o GAT)}: Removing the GAT components from GAMMA-Net led to increases in error metrics across all prediction horizons (MAE, RMSE, MAPE), underscoring the critical role of graph attention mechanisms in effectively modeling traffic data.
    
    \item \textbf{GAMMA-Net without Temporal Components (w/o Temporal GAMMA-Net)}: Omitting the temporal components slightly worsened the model's performance, indicating their importance in capturing time-dependent traffic patterns.
    
    \item \textbf{GAMMA-Net without Spatial Components (w/o Spatio GAMMA-Net)}: Removing the spatial components resulted in higher error metrics, highlighting their necessity for understanding spatial relationships within the data.
    
    \item \textbf{GAMMA-Net without Both Spatial and Temporal Components (w/o both GAMMA-Net)}: This configuration, which lacks both spatial and temporal components, exhibited the most significant performance drop, validating the synergistic importance of these elements in the model's architecture.
\end{itemize}

The fully integrated GAMMA-Net model consistently demonstrated superior performance across all measured metrics and prediction horizons compared to its ablated variants, emphasizing that both the spatial and temporal components of the GAT and Mamba are indispensable for achieving high accuracy and robustness in traffic forecasting tasks.

\textbf{Role of graph attention:}  
Eliminating the two GAT stages (w/o GAT) increases MAE by \(\,+2.4\,\%\) on METR-LA and \(+1.0\,\%\) on PEMS-BAY at the 15-min horizon, with the gap widening at 30 min and 60 min. Because the underlying road graph is fixed, these jumps cannot be attributed to missing connectivity; instead they indicate that edge re-weighting is indispensable for suppressing obsolete influences (e.g.\ closed ramps) and amplifying emergent ones (e.g.\ spill-back links).  Once this dynamic filtering is removed, the subsequent Mamba scans propagate stale signals and error accumulates.

\textbf{Role of the dual Mamba scans:}  
Omitting either the temporal (w/o Temporal GAMMA-Net) or the spatial (w/o Spatio GAMMA-Net) Mamba raises error modestly, yet removing both (w/o both GAMMA-Net) causes MAE to surge by \(+44\,\%\) (METR-LA) and \(+45\,\%\) (PEMS-BAY) at the one-hour horizon. This pattern confirms that the two axes are not interchangeable: Temporal Mamba compresses long sequences into memory-efficient hidden states, freeing later layers from the vanishing-gradient bottleneck; Spatial Mamba then disperses these context-rich signals over the graph in linear time, avoiding the parameter inflation of full graph convolutions. With only one axis present, half of this synergetic loop is broken; with both removed, the model reverts to a shallow GAT and performance collapses.

\textbf{Why the “GAT $\rightarrow$ Mamba$_\text{Temporal}$ $\rightarrow$ GAT $\rightarrow$ Mamba$_\text{Spatial}$” order?}  
The ablations also explain the formulation logic. Placing GAT before each Mamba scan supplies the state-space layers with up-to-date topological priors, while the ensuing Mamba outputs refresh node features that the next GAT pass can exploit. If GAT is kept but either Mamba axis is missing, node embeddings stop evolving along that dimension and the second GAT is left with stale features, leading to the incremental but measurable degradations seen in Table~\ref{tab:ablation}. Conversely, when both axes are active the information loop closes every block, yielding the lowest errors across all horizons.

\subsection{Visualization Study}
To further understand the GAMMA-Net model's performance, we conducted a visualization study focusing on three key aspects: the state transition matrices of the Mamba component, the attention weights from the GAT component, and the learned embeddings. These visualizations provide insights into how the model captures spatial and temporal dependencies in traffic data.

\subsubsection{SVD-Based Visualization and Analysis of State Transition Matrices in Mamba Component}
In the Mamba architecture, a selective state space model (SSM) efficiently processes long sequences~\cite{Mamba}. Under the SSM framework, the output at each time step is determined jointly by the current input and the state from the previous time step. The state transition operator A transfers the previous state \(h_{t-1}\) into the current state \(h_t\) through a linear transformation:
\begin{equation}
    h_t = A\, h_{t-1} + B\, x_t,
\end{equation}
where \(A\) is the state-transition matrix that governs how the state ``flows" or ``evolves"—similar to the hidden state propagation in RNNs—but in Mamba, this process is typically structured for efficient computation, \(B\) is the input-to-state matrix, and \(x_t\) is the input signal.

To understand and visualize the dynamics encoded by state transition matrices \( A \), we propose a model interpretability method based on singular value decomposition (SVD):
\begin{equation}
    A = U \Sigma V^\top,
\end{equation}
where \( U \) and \( V \) are orthogonal matrices, and \( \Sigma = \operatorname{diag}(\sigma_1, \sigma_2, \ldots, \sigma_n) \) contains the singular values.

The singular values \( \sigma_i \) offer crucial insights into the contributions of various components within the state transition matrix. By examining the magnitudes of these singular values, one can assess the system's stability—for example, if all singular values have magnitudes less than 1, the discrete system is considered stable. Moreover, the existence of dominant singular values (i.e., those with relatively larger magnitudes) indicates that certain components have a stronger influence on state evolution. This analysis not only infers the importance of different temporal dependencies but also provides a measure of the overall stability of the system.

\begin{figure*}
    \centering
    \subfigure[Spatial] {\includegraphics[width=0.45\textwidth]{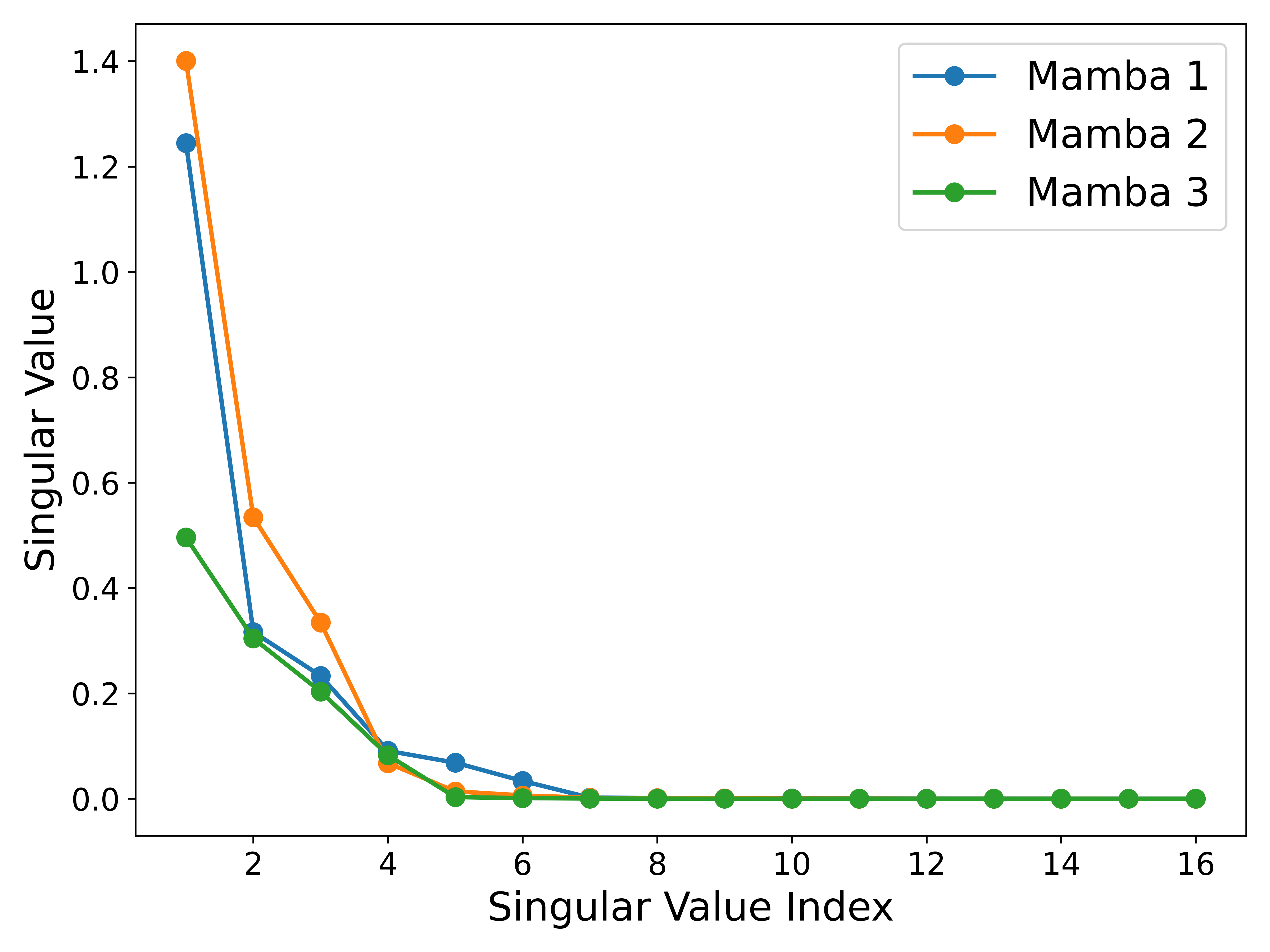}}
    \qquad
    \subfigure[Temporal] {\includegraphics[width=0.45\textwidth]{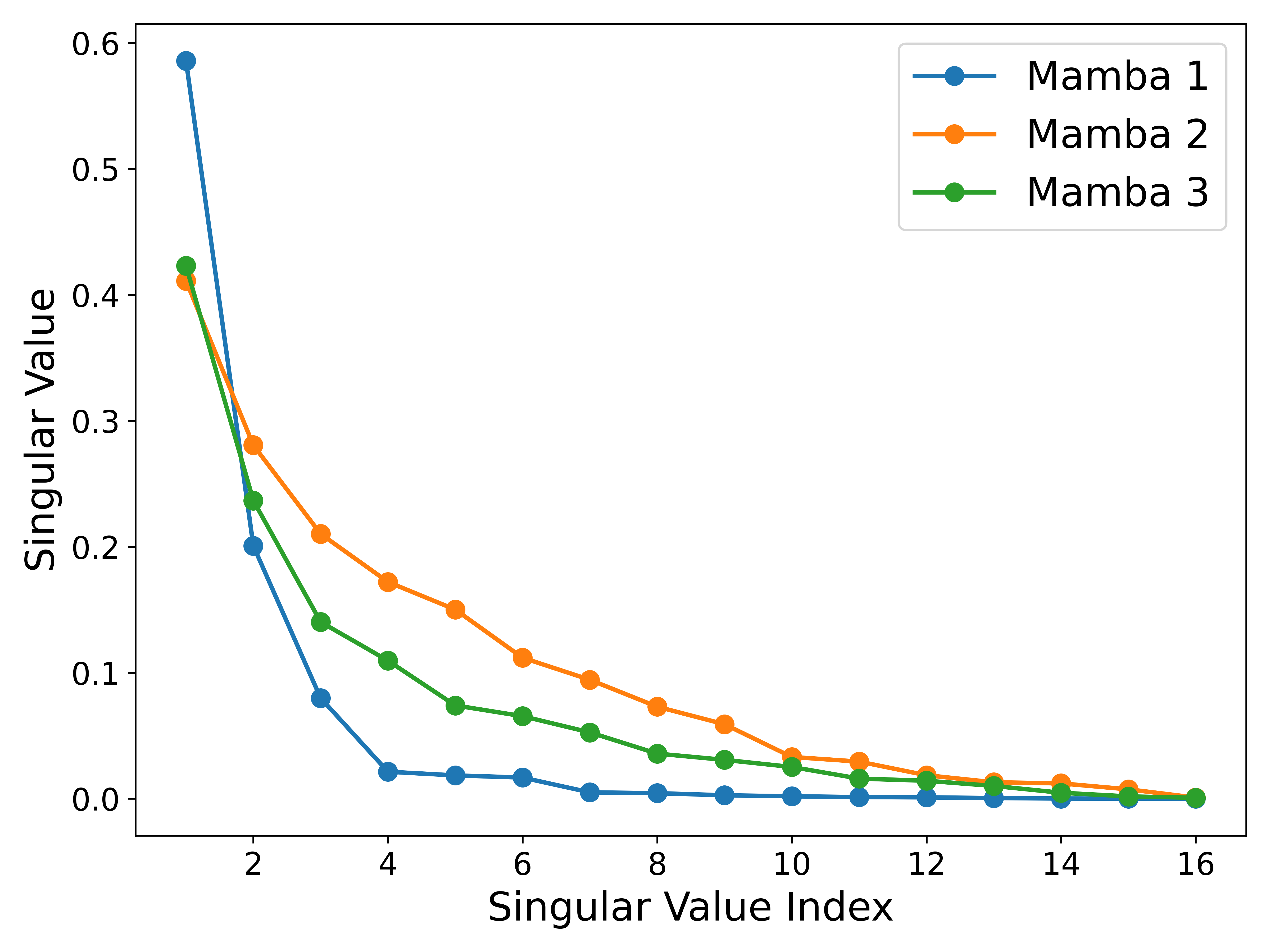}}
    \caption{SVD-based visualization of state transition matrices in the Mamba component of spatial and temporal GAMMA-Net trained by METR-LA.}
    \label{image:svd_full}
\end{figure*}

To illustrate this, we present the singular value distributions of the state transition matrices from both the spatial and temporal components of GAMMA-Net trained by METR-LA, as depicted in Figure~\ref{image:svd_full}. The left plot corresponds to the spatial component, and the right plot to the temporal component. The singular values are arranged in descending order to provide a clearer interpretation of their distribution. 

From the plots in Figure~\ref{image:svd_full}, we could observe that the spatial component exhibits a concentrated singular value distribution, suggesting that a few dominant components govern its state evolution. In contrast, the temporal component displays a more uniform distribution, implying that it captures a broader spectrum of temporal dependencies. This contrast aligns with our expectations: the spatial component is tailored to capture local dependencies, while the temporal component is designed to model long-range dynamics. Moreover, since all singular values are less than 1, both components are confirmed to be stable.

\subsubsection{Visualization of Graph Attention in GAMMA-Net}
To understand the latent graph structures that GAMMA-Net learns, we extract the edge indices and attention weights from the final spatial and temporal GAT layers-trained by METR-LA dataset-during inference. From these tensors we build two undirected, weighted graphs—one spatial, one temporal—by adding only those edges whose attention weight exceeds a threshold of 0.1, thereby filtering out weak, noisy connections.  We then apply the Louvain algorithm to detect communities in each graph.

\begin{figure*}
    \centering
    \subfigure[Spatial]{\includegraphics[width=0.45\textwidth]{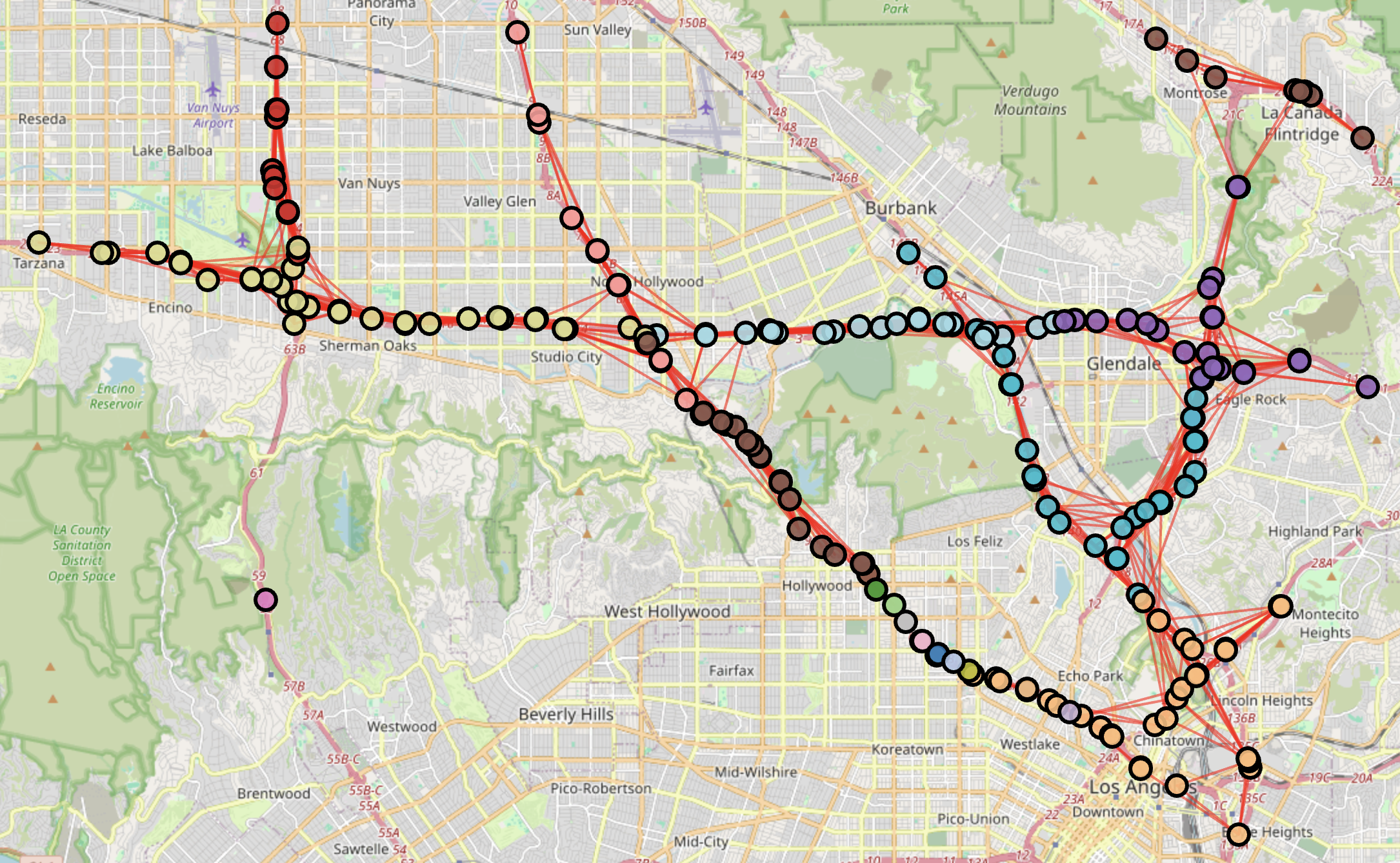}} 
    \qquad
    \subfigure[Temporal]{\includegraphics[width=0.45\textwidth]{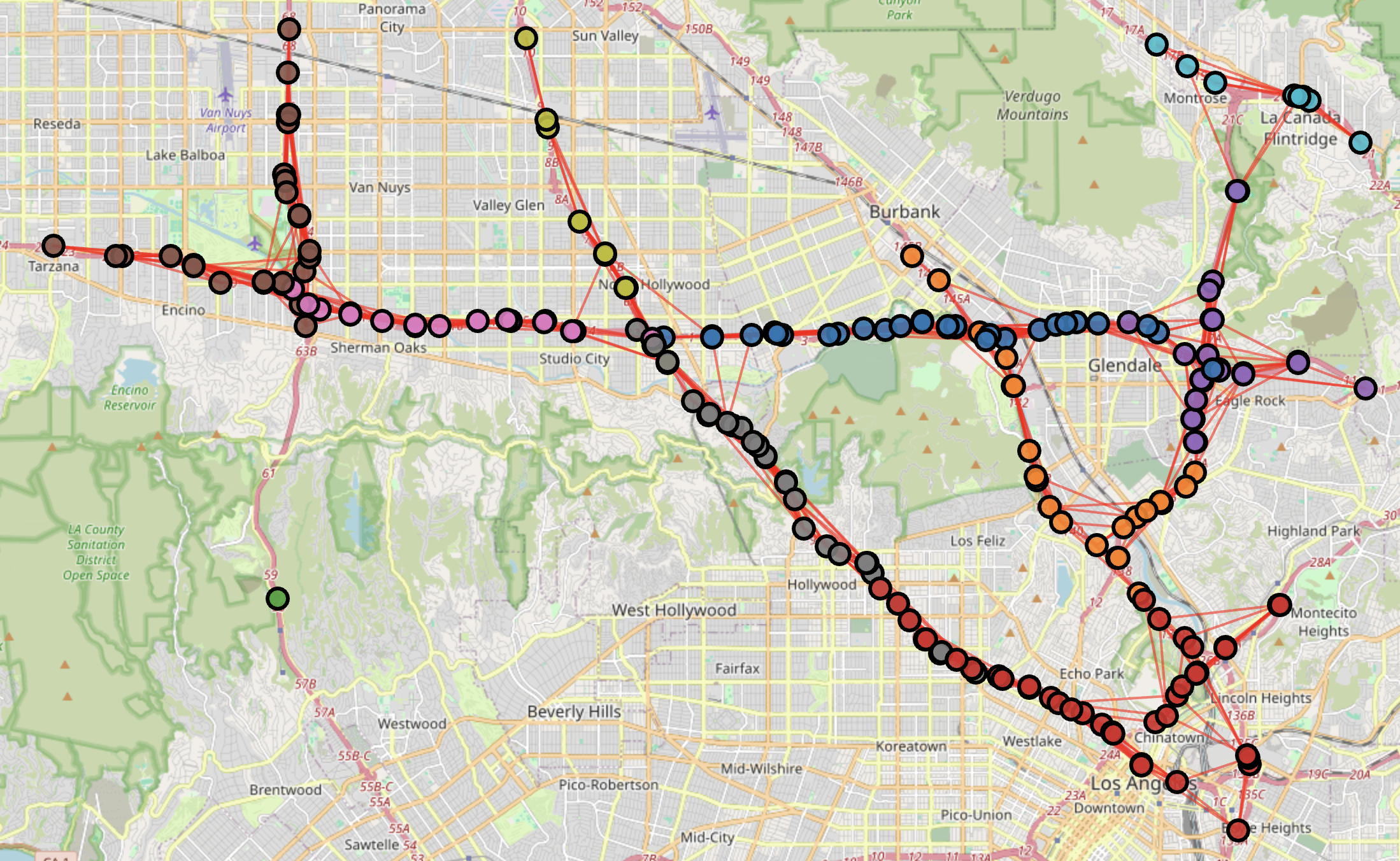}}
    \caption{Community detection based on spatial and temporal attention in the GAMMA-Net trained by METR-LA dataset. (a) Spatial community structure derived from attention weights and edge indices extracted from the final spatial attention layer. Each node is colored according to its spatial cluster, highlighting geographical patterns. (b) Temporal community structure based on temporal attention from the final temporal layer. Nodes with similar dynamic temporal dependencies are grouped into the same temporal cluster, regardless of spatial proximity. Both maps are overlaid on the Los Angeles region using real sensor locations.}
    \label{image:community}
\end{figure*}

As showed in Figure~\ref{image:community},  the spatial communities overlaid on the Los Angeles sensor network.  Each circle marks a sensor’s true latitude/longitude and is colored by its Louvain community in the spatial attention graph.  We see tight clusters along major freeways (e.g. east–west corridors through Glendale and the Hollywood Hills), and a distinct downtown grouping, illustrating that spatial attention faithfully captures geographic proximity. Temporal communities on the same map, derived from the temporal attention graph.  Here, sensors sharing similar time‐series behavior  are grouped—even if they are miles apart—revealing  functional connectivity.

\begin{figure*}
    \centering
    \subfigure[Spatial]{\includegraphics[width=0.45\textwidth]{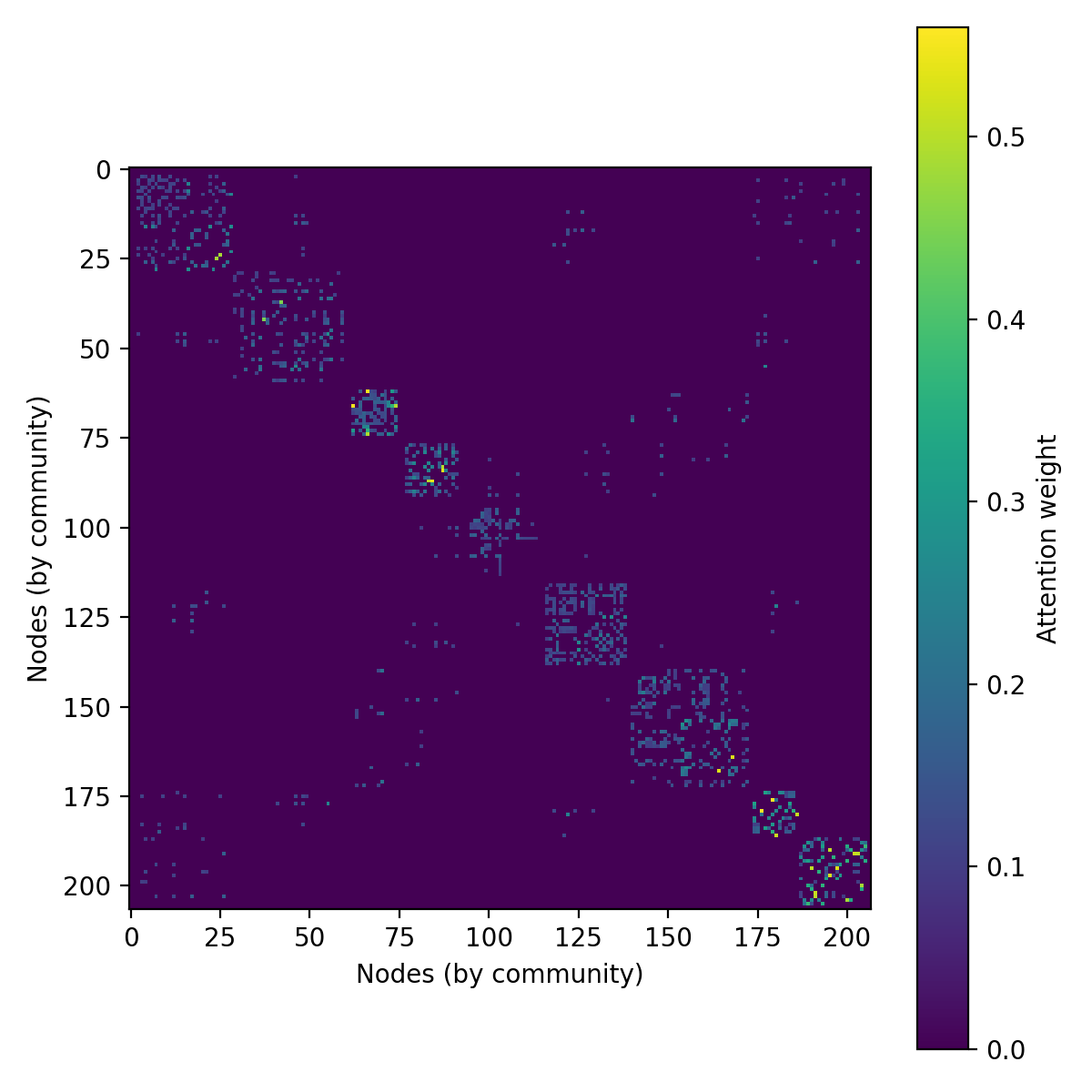}} \quad
    \subfigure[Temporal]{\includegraphics[width=0.45\textwidth]{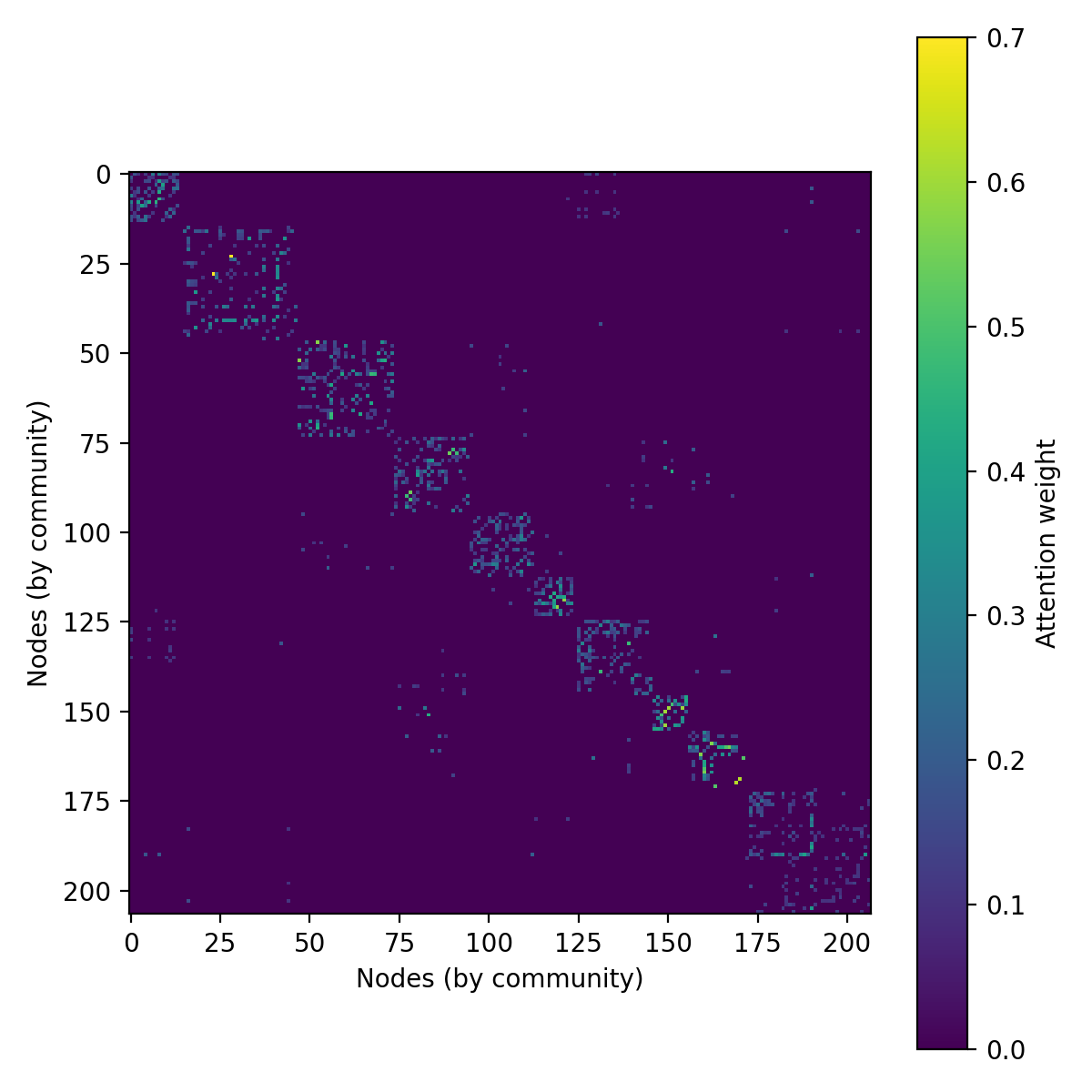}}
    \caption{Reordered adjacency matrices of the spatial and temporal attention graphs extracted from GAMMA-Net trained by METR-LA dataset. (a) shows the spatial attention adjacency matrix reordered by Louvain-detected community assignments. (b) presents the temporal attention adjacency matrix similarly reordered. Clear block-diagonal patterns reflect strong intra-community connectivity and sparse inter-community interactions, validating the effectiveness of attention mechanisms in capturing modular graph structure.}
    \label{image:adj}
\end{figure*}

As showed in Figure~\ref{image:adj},  the spatial attention adjacency matrix reordered by community membership.  Clear block-diagonal squares indicate strong intra-community weights; faint off-diagonal noise shows sparse links between communities.  The temporal attention matrix similarly reordered.  Here blocks are even denser and darker , reflecting more decisive intra-cluster temporal dependencies. The block-diagonal patterns in the heatmaps validate that both attention heads yield modular graphs with high within-community cohesion and low between-community connectivity, demonstrating the model's ability to adaptively learn meaningful graph topology along both spatial and temporal dimensions.

\subsubsection{Visualization of Learned Embedding}
\begin{figure*}
    \centering
    \includegraphics[width=0.7\textwidth]{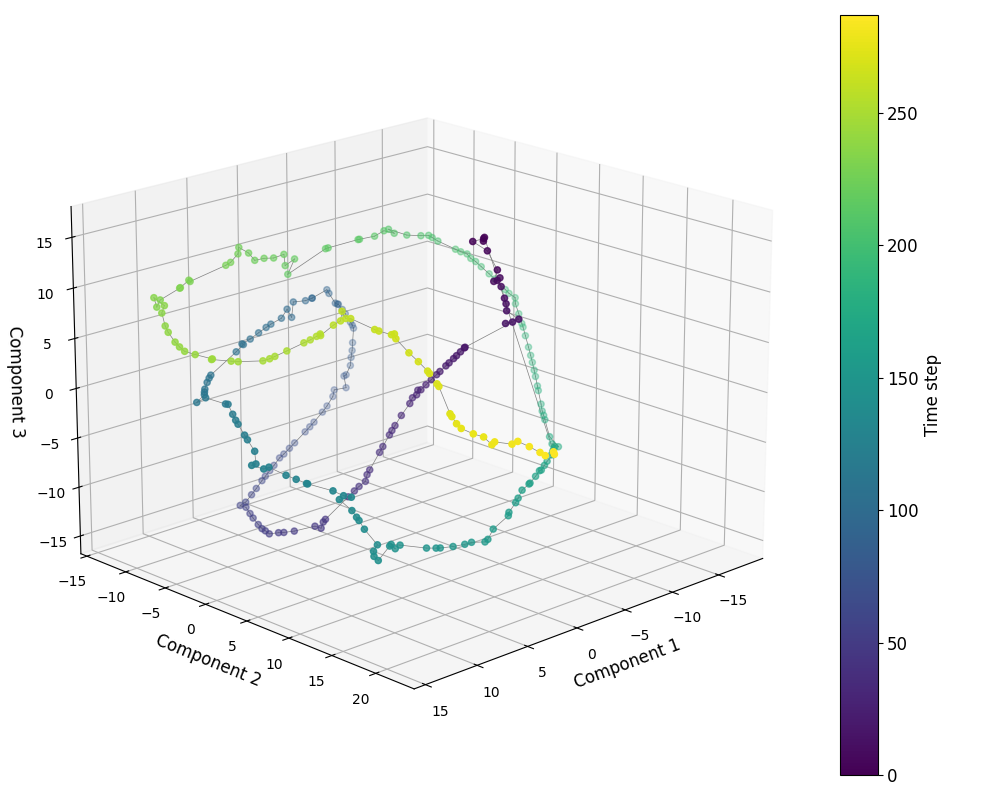}
    \caption{t-SNE visualization of the timestamp-of-day embeddings in GAMMA-Net. The reduced-dimensionality results are connected in time order (including the connection between the last and the first time point), and a gradient color scheme is applied to further represent the progression of time.}
    \label{image:embedding}
\end{figure*}

To further explore the learned embeddings in GAMMA-Net, we apply t-SNE~\cite{TSNE} for dimensionality reduction of the timestamp-of-day embeddings. The resulting visualization, presented in Figure~\ref{image:embedding}, illustrates how the model organizes these embeddings within the latent space. Notably, similar timestamps are clustered together while distinct timestamps are well separated—demonstrating that GAMMA-Net effectively captures underlying temporal patterns. This organized structure confirms the model's capacity to learn meaningful time representations, ultimately enhancing its predictive performance.

\subsection{Prediction Visualization}
\begin{figure*}
    \centering
    \includegraphics[width=\textwidth]{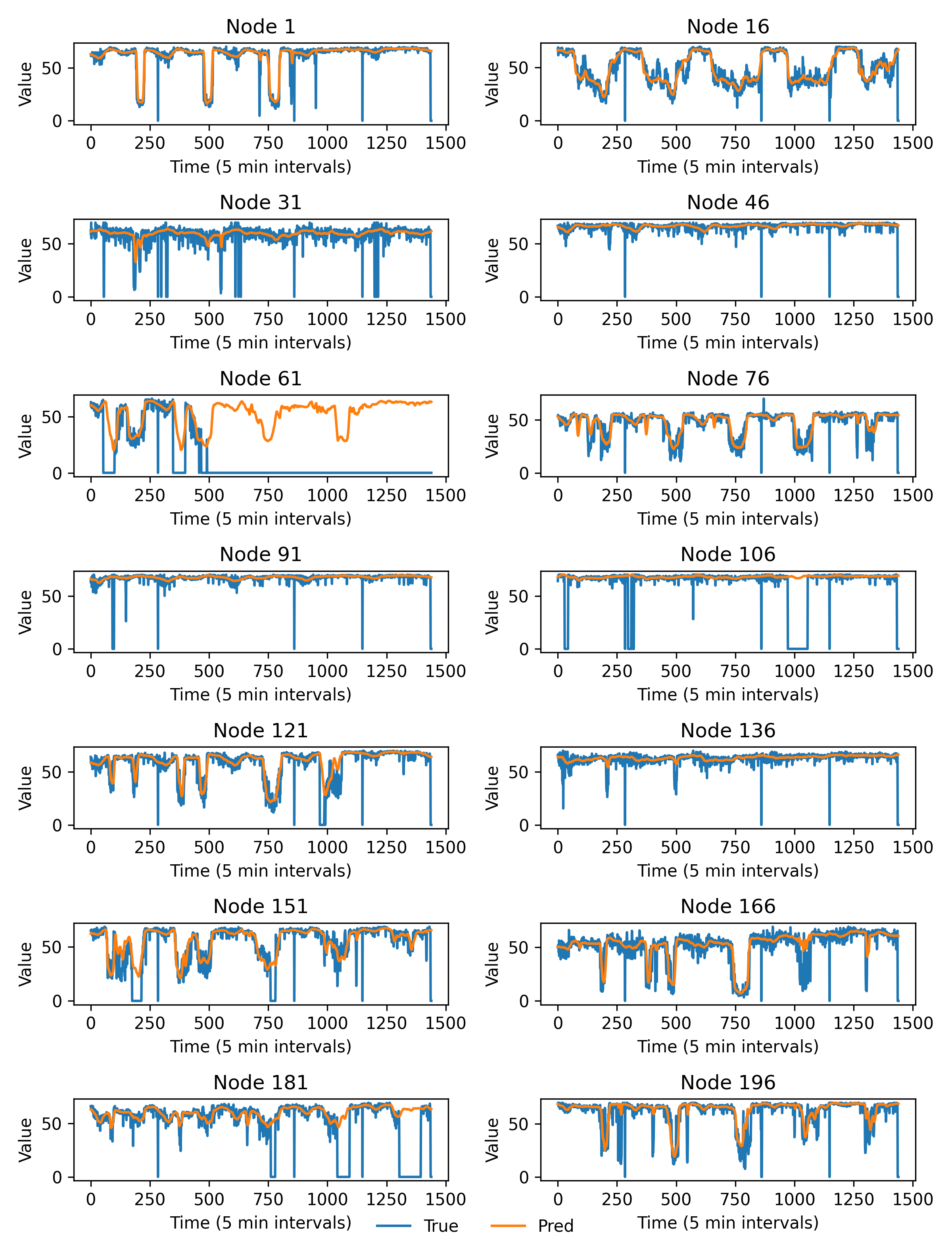}
    \caption{True vs. predicted traffic flow for 14 representative METR-LA sensor nodes. Each subplot displays approximately 1,440 time steps (5-minute intervals) of observed flow (blue) and the model's forecasted flow (orange) at a single node. The x-axis denotes time in 5-minute increments, and the y-axis shows traffic flow. Subplot titles indicate the sensor node index.}
    \label{fig:prediction_visualization}
\end{figure*}

To demonstrate the forecasting capability and robustness of GAMMA-Net on the METR-LA dataset, we uniformly sample 14 sensor nodes at equal index intervals—covering diverse spatial regions and traffic regimes—and compare their true and predicted traffic volumes. Figure~\ref{fig:prediction_visualization} overlays the ground-truth series (blue)—which include intermittent zeros due to sensor dropouts treated as missing values—and the model's continuous forecasts (orange) over the full test period (approximately 1,440 steps at 5-minute intervals).

The proposed GAMMA-Net accurately captures the diurnal rush-hour patterns—both the sharp morning drops and gradual recoveries—while remaining remarkably insensitive to isolated zeros, and the model closely reproduces both peak magnitudes and off-peak troughs. At more stable locations, predictions stay uniformly flat outside rush hours, with minimal deviation from the true volumes.

Despite frequent zero readings (e.g., Node 61 and Node 106) caused by sensor malfunctions or data gaps, GAMMA-Net’s forecasts remain smooth and plausible, implicitly “inpainting” missing-value outliers by leveraging both spatial neighbors and temporal context. This demonstrates the model’s robustness: it does not overreact to spurious zeros but instead maintains consistent predictive performance across varying data quality conditions.

\begin{figure*}
    \centering
    \includegraphics[width=\textwidth]{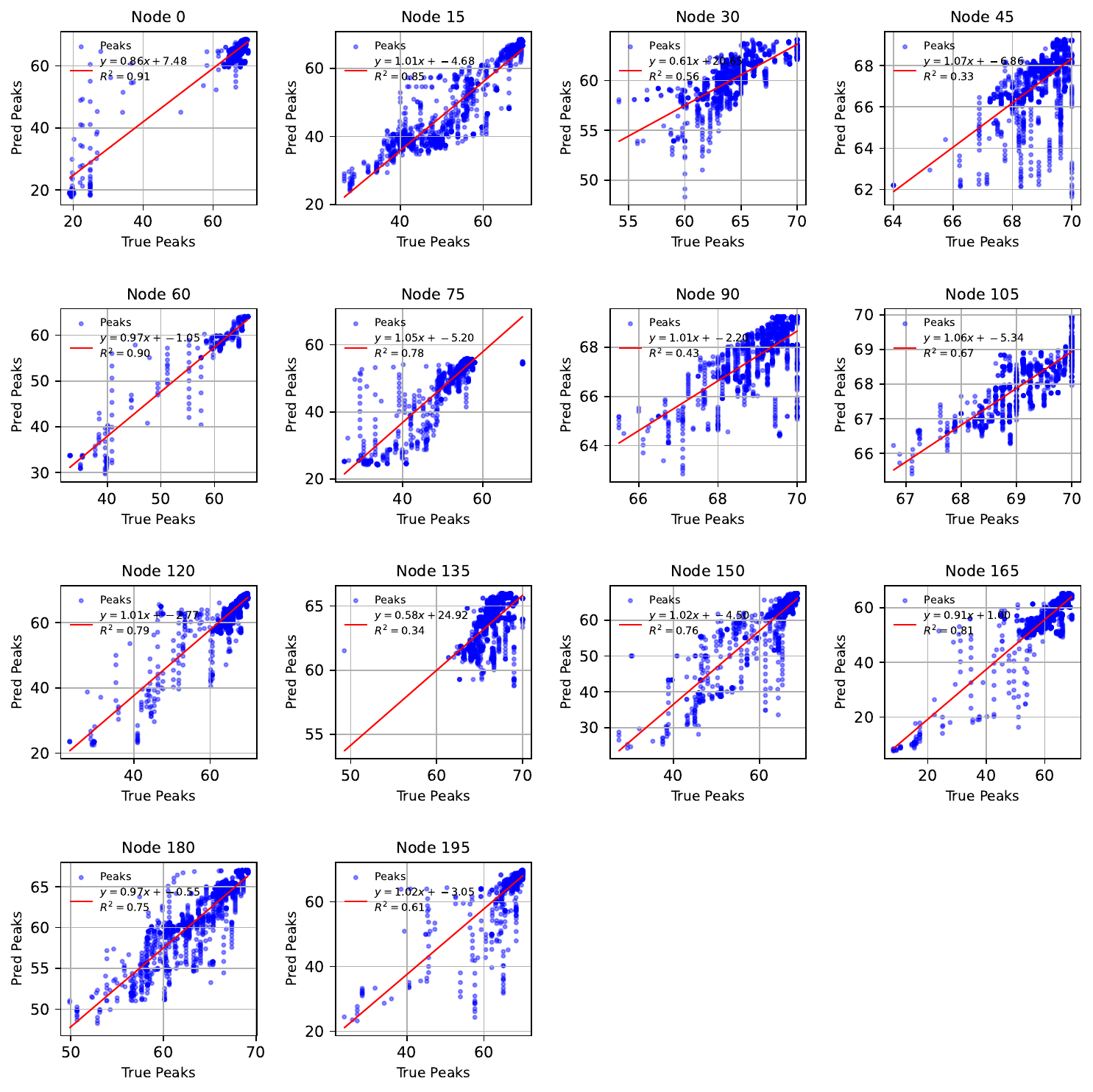}
    \caption{Scatter plots of predicted versus true peak values across selected nodes. Each subplot corresponds to one node (sampled every 15 nodes), with peak values extracted using a sliding window of size 12. Red lines indicate fitted linear regression models, with corresponding regression equations and coefficients of determination ($R^2$) displayed.}
    \label{fig:peak_node_regression}
\end{figure*}

In order to quantify GAMMA-Net's ability to reproduce extreme events, we extracted peak values from both the true and predicted time series using a sliding window of length 12 at fifteen spatially distributed sensor nodes. For each node, as showed in Figure~\ref{fig:peak_node_regression}, we then plotted the true peaks against the predicted peaks and fitted an ordinary least squares line, reporting both the regression equation $y = a x + b$ and the coefficient of determination $R^2$ in each subplot. Across all nodes, the estimated slopes $a_i$ cluster tightly around 1.0, indicating that GAMMA-Net neither systematically over- nor under-scales peak magnitudes, while the intercepts $b_i$ remain modest. The majority of nodes achieve high explanatory power ($R^2_i>0.7$), with the best fits at Node 0 ($R^2=0.91$) and Node 60 ($R^2=0.90$), underscoring the model's strong performance in relatively stable traffic regimes.

\label{sec:discussion}
\section{Discussion}
\subsection{Strengths and Limitations}
GAMMA-Net exhibits several notable strengths that contribute to its superior performance in traffic forecasting. First, the model enhances prediction accuracy by dynamically weighting the influence of individual nodes and efficiently managing temporal dependencies. This dual approach leads to substantial reductions in forecasting errors. Moreover, the state space component enables parallel processing along the temporal axis, which significantly improves computational efficiency and makes the model well-suited for real-time applications compared to traditional recurrent architectures.

Another advantage of GAMMA-Net is its scalability. By leveraging the inherent sparsity of traffic networks, the model maintains high computational efficiency even as the size of the network increases, thus supporting large-scale deployments and complex urban environments.

Despite these strengths, the study also uncovers several limitations. For instance, the reliance on a pre-defined graph structure may restrict the model's adaptability in scenarios where network topologies change rapidly. Additionally, while GAMMA-Net achieves impressive predictive performance, its interpretability in complex, real-world settings remains limited—a common challenge among deep learning models applied to spatiotemporal data.

\subsection{Reasons for Performance Improvement}
The improved performance of GAMMA-Net arises from its ability to effectively integrate adaptive spatial modeling with efficient long-term temporal dynamics. Unlike traditional models that rely on pre-defined or static graph structures, our approach utilizes Graph Attention Networks to dynamically assign weights to neighboring nodes based on real-time traffic conditions. This dynamic weighting mechanism enables the model to capture subtle spatial dependencies and adapt to varying traffic patterns. At the same time, the incorporation of the selective state-space module (Mamba) allows the model to efficiently preserve and update relevant temporal information without encountering issues such as vanishing gradients or excessive computational overhead that are common in recurrent architectures. The combination of these two strategies not only captures both short-term fluctuations and long-range trends in traffic data but also results in significant improvements in forecasting accuracy when compared to baseline methods.

\subsection{Computational Efficiency Advantages}
In addition to the accuracy benefits, GAMMA-Net demonstrates clear computational efficiency advantages that are critical for real-time applications. The temporal component based on the state-space formulation permits parallel processing along the time axis, which contrasts with the sequential operations of traditional recurrent models. This parallelism reduces both training and inference times, making the model more practical for scenarios where rapid response is essential. Moreover, the graph attention mechanism focuses computations on the most relevant neighboring nodes rather than processing an entirely dense graph, thereby taking advantage of the inherent sparsity in traffic networks. By leveraging this sparsity, GAMMA-Net minimizes unnecessary computations and allocates resources more efficiently, resulting in a model that is not only faster but also more scalable than many existing approaches.

\section{Conclusion}
\label{sec:conclusion}

In this study, we introduced the novel \textbf{GAMMA-Net} model, which synergistically integrates Graph Attention Networks (GAT) with Selective State Space Models (Mamba) to address the challenging task of spatiotemporal traffic forecasting. Extensive experiments on benchmark datasets, including METR-LA and PEMS-BAY, demonstrate that GAMMA-Net outperforms existing state-of-the-art methods across multiple prediction horizons. This performance gain is primarily attributed to the model’s dual mechanism: its ability to dynamically capture spatial dependencies through attention-based graph processing, and to efficiently model long-term temporal dynamics via a robust state space formulation.

Comprehensive ablation studies affirm the individual and collective contributions of the spatial and temporal components. The results indicate that while each module contributes uniquely to the overall performance, their integration is essential for achieving optimal accuracy in forecasting. These findings underscore the importance of jointly addressing spatial heterogeneity and temporal evolution in complex traffic data.

Future research directions include the development of dynamic graph construction techniques that can adaptively update network structures in real time, thereby enhancing model flexibility and responsiveness. Additionally, efforts will be directed toward improving the interpretability of the framework through advanced visualization and analysis techniques, which will be invaluable for its integration into practical traffic management systems. Moreover, given the general applicability of the proposed approach, we plan to extend GAMMA-Net to other domains characterized by graph-based spatiotemporal data, such as weather forecasting and dynamic network analysis.

In summary, the GAMMA-Net model not only establishes a new benchmark in traffic forecasting but also lays a robust foundation for future research in spatiotemporal predictive analytics. The innovative integration of graph attention mechanisms with selective state space modeling paves the way for the development of smarter and more adaptive traffic management systems.

\section*{Acknowledgments}\label{sec:acknowledgments}
This work is supported by the Scientific and Technological Research Program of the Chongqing Education Commission (KJQN202201142), the Natural Science Foundation of Chongqing (2024NSCQ-MSX1731), the Chongqing University of Technology Research and Innovation Team Cultivation Program (2023TDZ012), the Chongqing Municipal Key Project for Technology Innovation and Application Development (CSTB2024TIAD-KPX0042), and the National Natural Science Foundation of P.R. China (61173184).

\section*{Declarations}\label{sec:Declarations}
\begin{itemize}
    \item \textbf{Competing Interests} All the authors declare that they have no conflict of interest.
    \item \textbf{Authors contribution statement} Conceptualization, Methodology, Writing-original draft: [First Author]; Visualization, Validation, Writing-review \& editing: [Second Author]; Supervision, Funding acquisition: [Third and Fourth Authors].
    \item \textbf{Data Availability and Access} The preprocessed datasets—including traffic flow data and corresponding graph structures—used in this study are available at: \url{https://drive.google.com/drive/folders/1fV0IAUsjHeNeoqq1kdyqn-6XYxm7P566?usp=sharing}. The code for this study is available at \url{ https://github.com/hdy6438/GAMMA-Net}
\end{itemize}

\bibliographystyle{unsrt}
\bibliography{references}  

@inproceedings{STAEformer,
  title={STAEformer: Spatio-temporal adaptive embedding makes vanilla transformer sota for traffic forecasting},
  author={Liu, Hangchen and Dong, Zheng and Jiang, Renhe and Deng, Jiewen and Deng, Jinliang and Chen, Quanjun and Song, Xuan},
  booktitle={Proceedings of the 32nd ACM international conference on information and knowledge management},
  pages={4125--4129},
  year={2023}
}

@article{GAT,
  title={Graph attention networks},
  author={Velickovic, Petar and Cucurull, Guillem and Casanova, Arantxa and Romero, Adriana and Lio, Pietro and Bengio, Yoshua and others},
  journal={stat},
  volume={1050},
  number={20},
  pages={10--48550},
  year={2017}
}

@article{Mamba,
  title={Mamba: Linear-time sequence modeling with selective state spaces},
  author={Gu, Albert and Dao, Tri},
  journal={arXiv preprint arXiv:2312.00752},
  year={2023}
}

@article{GWNet,
  title={Graph wavenet for deep spatial-temporal graph modeling},
  author={Wu, Zonghan and Pan, Shirui and Long, Guodong and Jiang, Jing and Zhang, Chengqi},
  journal={arXiv preprint arXiv:1906.00121},
  year={2019}
}

@article{DCRNN,
  title={Diffusion convolutional recurrent neural network: Data-driven traffic forecasting},
  author={Li, Yaguang and Yu, Rose and Shahabi, Cyrus and Liu, Yan},
  journal={arXiv preprint arXiv:1707.01926},
  year={2017}
}

@article{AGCRN,
  title={Adaptive graph convolutional recurrent network for traffic forecasting},
  author={Bai, Lei and Yao, Lina and Li, Can and Wang, Xianzhi and Wang, Can},
  journal={Advances in neural information processing systems},
  volume={33},
  pages={17804--17815},
  year={2020}
}

@article{STGCN,
  title={Spatio-temporal graph convolutional networks: A deep learning framework for traffic forecasting},
  author={Yu, Bing and Yin, Haoteng and Zhu, Zhanxing},
  journal={arXiv preprint arXiv:1709.04875},
  year={2017}
}

@article{GTS,
  title={Discrete graph structure learning for forecasting multiple time series},
  author={Shang, Chao and Chen, Jie and Bi, Jinbo},
  journal={arXiv preprint arXiv:2101.06861},
  year={2021}
}

@inproceedings{MTGNN,
  title={Connecting the dots: Multivariate time series forecasting with graph neural networks},
  author={Wu, Zonghan and Pan, Shirui and Long, Guodong and Jiang, Jing and Chang, Xiaojun and Zhang, Chengqi},
  booktitle={Proceedings of the 26th ACM SIGKDD international conference on knowledge discovery \& data mining},
  pages={753--763},
  year={2020}
}

@inproceedings{STNorm,
  title={St-norm: Spatial and temporal normalization for multi-variate time series forecasting},
  author={Deng, Jinliang and Chen, Xiusi and Jiang, Renhe and Song, Xuan and Tsang, Ivor W},
  booktitle={Proceedings of the 27th ACM SIGKDD conference on knowledge discovery \& data mining},
  pages={269--278},
  year={2021}
}

@inproceedings{GMAN,
  title={Gman: A graph multi-attention network for traffic prediction},
  author={Zheng, Chuanpan and Fan, Xiaoliang and Wang, Cheng and Qi, Jianzhong},
  booktitle={Proceedings of the AAAI conference on artificial intelligence},
  volume={34},
  number={01},
  pages={1234--1241},
  year={2020}
}

@inproceedings{PDFormer,
  title={Pdformer: Propagation delay-aware dynamic long-range transformer for traffic flow prediction},
  author={Jiang, Jiawei and Han, Chengkai and Zhao, Wayne Xin and Wang, Jingyuan},
  booktitle={Proceedings of the AAAI conference on artificial intelligence},
  volume={37},
  number={4},
  pages={4365--4373},
  year={2023}
}

@inproceedings{STID,
  title={Spatial-temporal identity: A simple yet effective baseline for multivariate time series forecasting},
  author={Shao, Zezhi and Zhang, Zhao and Wang, Fei and Wei, Wei and Xu, Yongjun},
  booktitle={Proceedings of the 31st ACM International Conference on Information \& Knowledge Management},
  pages={4454--4458},
  year={2022}
}

@inproceedings{STSGCN,
  title={Spatial-temporal synchronous graph convolutional networks: A new framework for spatial-temporal network data forecasting},
  author={Song, Chao and Lin, Youfang and Guo, Shengnan and Wan, Huaiyu},
  booktitle={Proceedings of the AAAI conference on artificial intelligence},
  volume={34},
  number={01},
  pages={914--921},
  year={2020}
}

@article{Vaswani2017,
  title={Attention is all you need},
  author={Vaswani, A},
  journal={Advances in Neural Information Processing Systems},
  year={2017}
}

@article{Kitaev2020Reformer,
  title={Reformer: The efficient transformer},
  author={Kitaev, Nikita and Kaiser, {\L}ukasz and Levskaya, Anselm},
  journal={arXiv preprint arXiv:2001.04451},
  year={2020}
}

@article{GCN,
  title={Graph convolutional networks: a comprehensive review},
  author={Zhang, Si and Tong, Hanghang and Xu, Jiejun and Maciejewski, Ross},
  journal={Computational Social Networks},
  volume={6},
  number={1},
  pages={1--23},
  year={2019},
  publisher={Springer}
}

@article{TSNE,
  title={Visualizing data using t-SNE.},
  author={Van der Maaten, Laurens and Hinton, Geoffrey},
  journal={Journal of machine learning research},
  volume={9},
  number={11},
  year={2008}
}

@article{lablack2023spatio,
  title={Spatio-temporal graph mixformer for traffic forecasting},
  author={Lablack, Mourad and Shen, Yanming},
  journal={Expert systems with applications},
  volume={228},
  pages={120281},
  year={2023},
  publisher={Elsevier}
}

@article{afandizadeh2024deep,
  title={Deep Learning Algorithms for Traffic Forecasting: A Comprehensive Review and Comparison with Classical Ones},
  author={Afandizadeh, Shahriar and Abdolahi, Saeid and Mirzahossein, Hamid},
  journal={Journal of Advanced Transportation},
  volume={2024},
  number={1},
  pages={9981657},
  year={2024},
  publisher={Wiley Online Library}
}

@inproceedings{yuan2024st,
  title={ST-Mamba: Spatial-Temporal Mamba for Traffic Flow Estimation Recovery using Limited Data},
  author={Yuan, Doncheng and Xue, Jianzhe and Su, Jinshan and Xu, Wenchao and Zhou, Haibo},
  booktitle={2024 IEEE/CIC International Conference on Communications in China (ICCC)},
  pages={1928--1933},
  year={2024},
  organization={IEEE}
}

@inproceedings{he2025decomposed,
  title={Decomposed Spatio-Temporal Mamba for Long-Term Traffic Prediction},
  author={He, Sicheng and Ji, Junzhong and Lei, Minglong},
  booktitle={Proceedings of the AAAI Conference on Artificial Intelligence},
  volume={39},
  number={11},
  pages={11772--11780},
  year={2025}
}

@article{shao2024st,
  title={ST-MambaSync: The Complement of Mamba and Transformers for Spatial-Temporal in Traffic Flow Prediction},
  author={Shao, Zhiqi and Yao, Xusheng and Wang, Ze and Gao, Junbin},
  journal={arXiv preprint arXiv:2404.15899},
  year={2024}
}

@article{meric2025ms,
  title={ms-Mamba: Multi-scale Mamba for Time-Series Forecasting},
  author={Meric Karadag, Yusuf and Kalkan, Sinan and Gursel Dino, Ipek},
  journal={arXiv e-prints},
  pages={arXiv--2504},
  year={2025}
}

@article{li2024stg,
  title={Stg-mamba: Spatial-temporal graph learning via selective state space model},
  author={Li, Lincan and Wang, Hanchen and Zhang, Wenjie and Coster, Adelle},
  journal={arXiv preprint arXiv:2403.12418},
  year={2024}
}

@article{choi2024spot,
  title={Spot-mamba: Learning long-range dependency on spatio-temporal graphs with selective state spaces},
  author={Choi, Jinhyeok and Kim, Heehyeon and An, Minhyeong and Whang, Joyce Jiyoung},
  journal={arXiv preprint arXiv:2406.11244},
  year={2024}
}

@article{gao2023spatial,
  title={Spatial-temporal-decoupled masked pre-training for spatiotemporal forecasting},
  author={Gao, Haotian and Jiang, Renhe and Dong, Zheng and Deng, Jinliang and Ma, Yuxin and Song, Xuan},
  journal={arXiv preprint arXiv:2312.00516},
  year={2023}
}

\end{document}